\pdfoutput=1

\documentclass[11pt]{article}

\usepackage[final]{acl}
\usepackage{natbib}
\usepackage{amsmath}
\usepackage{amsfonts}
\usepackage{booktabs}
\usepackage{multirow}
\usepackage{placeins}
\usepackage{enumitem}

\usepackage{tikz}
\usepackage{tikz-cd}
\usetikzlibrary{arrows.meta, positioning}

\tikzset{
  data/.style={rounded corners, draw, fill=blue!5},   
  metrics/.style={rounded corners, draw, fill=orange!15},
  process/.style={rounded corners, draw},             
}

\newcommand{\RR}{\mathbb{R}}

\newcommand{\crbr}[1]{\left \{ #1 \right \} }

\newcommand{\noi}{\noindent}
\renewcommand*{\do}[1]{%
\expandafter\newcommand \csname c#1\endcsname{\mathcal{#1}}}
\docsvlist{A,B,C,D,E,F,G,H,I,J,K,L,M,N,O,P,Q,R, RP, DP, S,T,U,V,W,X,Y,Z, G W}

\usepackage{times}
\usepackage{latexsym}
\usepackage{tabularray} 

\usepackage[T1]{fontenc}

\usepackage[utf8]{inputenc}

\usepackage{microtype}

\usepackage{inconsolata}

\usepackage{graphicx}
\usepackage{subcaption}

\title{Mechanistic Decomposition of Sentence Representations}

\author{
Matthieu Tehenan$^{1}$,
Vikram Natarajan$^{2}$,
Jonathan Michala$^{3}$,
Milton Lin$^{4}$,
Juri Opitz$^{5}$ \\
\\
$^1$University of Cambridge \quad
$^2$Independent \quad
$^3$University of Southern California \\
$^4$Johns Hopkins University \quad
$^5$University of Zürich \\
\\
\texttt{mm2833@cam.ac.uk}
}

\begin{document}

\maketitle
\begin{abstract}
Sentence embeddings are central to modern NLP and AI systems, yet little is known about their internal structure. While we can compare these embeddings using measures such as cosine similarity, the contributing features are not human-interpretable, and the content of an embedding seems untraceable, as it is masked by complex neural transformations and a final pooling operation that combines individual token embeddings. To alleviate this issue, we propose a new method to mechanistically decompose sentence embeddings into interpretable components, by using dictionary learning on token-level representations. We analyze how pooling compresses these features into sentence representations, and assess the latent features that reside in a sentence embedding. This bridges token-level mechanistic interpretability with sentence-level analysis, making for more transparent and controllable representations. In our studies, we obtain several interesting insights into the inner workings of sentence embedding spaces, for instance, that many semantic and syntactic aspects are linearly encoded in the embeddings.
\end{abstract}

\begin{tblr}{colspec = {Q[c,m]|X[l,m]}, stretch = 0}
    \includegraphics[width=1.2em, keepaspectratio]{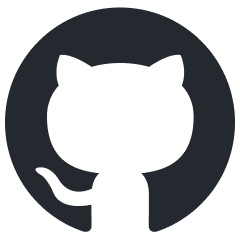}
     & \href{https://github.com/matthieu-perso/mechanistic_decomposition_sentences}{{\textsf{\small mechanistic-decomposition-sentences}}} \\
\end{tblr}

\section{Introduction}

\begin{figure}[t]
    \centering
    \includegraphics[width=\columnwidth]{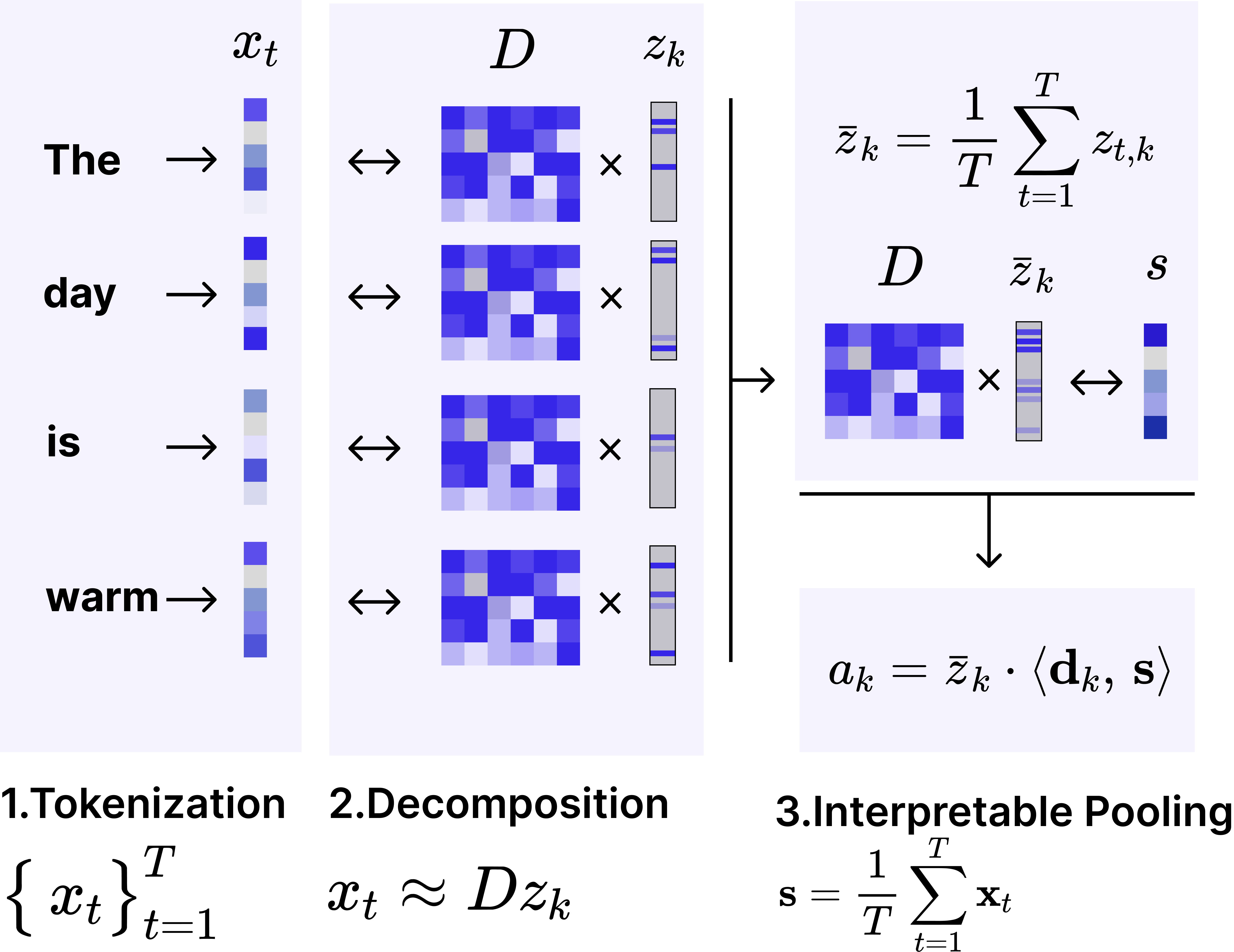}
    \caption{\small Overview of our method. (1) A sentence is tokenized into individual token embeddings $x_t$. (2) Each token embedding is decomposed into a set of interpretable latent features using dictionary learning $Dz_k$. (3) The decomposed features are aggregated via mean pooling of the sparse codes. This yields an interpretable decomposition of the pooled sentence vector $s$.
} 
    \label{fig:overview}
\end{figure}

Sentence embeddings are high-dimensional, distributed vector representations of text. As such, they are central to many of today's NLP tasks, ranging from similarity and retrieval applications to classification, clustering and paraphrasing applications \citep{lewis2020retrieval, gao2023retrieval, barnabo-etal-2023-supervised, han2023comprehensive}. 

However, embedding interpretability remains a challenge: The information encoded in contextualized representations presents itself as highly entangled, making it unclear which specific elements contribute to a given representation. While this is true also for word embeddings \citep{bommasani-etal-2020-interpreting}, the issue is particularly acute in the case of sentence embeddings \citep{opitz_interpretable_2025}. Sentence representations are typically derived from token-level contextualized encodings and subsequently pooled, which aggregates information across multiple tokens \citep{reimers_sentence-bert_2019, gao2021simcse, kashyap2023comprehensive}. The result is a dense structure, where linguistic features, contextual nuances, and superficial cues are commingled and cannot be easily separated. This entanglement complicates both the understanding of how such embeddings are applied to downstream tasks (e.g., sentence similarity), and more fundamentally, the ability to decompose and explain the representations themselves. 


On one hand, some first efforts have been made to address sentence embedding interpretability, e.g., by distilling interpretable metrics \citep{opitz_sbert_2022, benara2024crafting}, or assessing saliency of input tokens \citep{moeller-etal-2023-attribution, moeller-etal-2024-approximate, vasileiou-eberle-2024-explaining}. However, these methods suffer from drawbacks: E.g., metric distillation requires subjective design of metrics and retro-fitting of the model; token input saliency is extremely compute-heavy and backtracking to the input layer is not practical, hence an arbitrary intermediate layer has to be selected.

Instead, we propose to adopt the lens of \textit{mechanistic interpretability}. The overarching goal of this paradigm is to uncover the internal causal and structural mechanisms of models, essentially reverse-engineering the underlying computations to identify how specific components contribute to an output. Recent work in this direction has shown success in analyzing token-level representations in text generation models, offering causal results into how information flows and is transformed within the network \citep{bereska_mechanistic_2024, bricken2023monosemanticity}. These advances point to the possibility of applying causal interpretability methods not just at the token level, but also at the level of aggregated sentence embeddings. If we can trace how individual token representations are transformed and pooled into a sentence embedding, we can potentially construct interpretable causal mappings from inputs to representations and downstream behaviors. 

As described in \autoref{fig:overview}, our paper thus marks the first step towards mechanistic interpretability of sentence embeddings, 
bridging the gap between mechanistic interpretability and sentence-level representation analysis. Our main contributions are three-fold: 

\begin{enumerate}
  \setlength{\itemsep}{0pt}
  \item We propose a mechanistic framework for decomposing token embeddings into latent, interpretable components and study how mean pooling integrates these components into the final sentence representation.
  \item We apply supervised dictionary learning to extract latent components from the underlying token embeddings of these sentence representations and demonstrate their interpretability.
  \item We investigate the effects of mean pooling on these components, identifying which are preserved in the sentence representation and which are lost. As one important insight from this study, we show that some semantic aspects reside isolated in linear subspaces.
\end{enumerate}

\section{Background and Related Work}
\label{sec:background}


\subsection{Sentence Embeddings}

Sentence embeddings aim to generate representations that encapsulate the semantic meaning of sentences  \citep{wieting2015towards, reimers2019sentence, gao2021simcse}. Today, methods typically rely on contextualized token embeddings from transformer models \citep{li2020sentence, kashyap2023comprehensive}. 
To construct a single embedding from the set of token embeddings, various pooling methods have been proposed, where \textit{mean pooling}\footnote{The token embeddings are averaged across dimensions.} is currently the technique most widely employed, both due to its simplicity and its performance on downstream tasks \citep{arora2017simple, acs_subword_2021}. Since sentences contain semantic properties that are not contained in individual words, the underlying token-encoder is tuned on contrastive learning tasks like entailment and textual similarity, reshaping contextual token representations \citep{gao2021simcse, liu2020survey, kashyap2023comprehensive}. In this matter, our paper makes a significant contribution in understanding this common pooling mechanism, showing a way to study how individual semantics are aggregated from a semantic dictionary.

\subsection{Sentence Embedding Interpretability}

The interpretability of sentence embeddings has mostly been investigated in ways that are non-mechanistics based. For instance, one common approach involves decomposing these dense representations into latent, interpretable features derived from the full sentence, e.g., either by distilling custom metrics \citep{opitz_sbert_2022}, answering questions about the sentence with LLMs \citep{benara2024crafting}, or incorporating compositional operators to structure semantic relationships \citep{huang_bridging_2023}. The need for tailoring measures or other hyperparameters limits their scalability and generalizability across tasks and domains \citep{opitz_interpretable_2025}. Other drawbacks are incurred by input-saliency interpreatbility methods like attribution \citep{moeller-etal-2023-attribution, moeller-etal-2024-approximate}, or layer-wise backpropagation \citep{vasileiou-eberle-2024-explaining}. These are only aimed to explain the similarity itself (as opposed to the content of embeddings), and they are extremely compute-heavy. Embeddings have also been interpreted through behavior on tailored datasets, but this requires specialized linguistic annotation efforts \citep{nastase-merlo-2024-tracking, fodorchallenge}.  In our piece, we provide a complementary viewpoint, and make a first effort to apply concepts from mechanistic interpretability to better understand the semantic content of sentence embeddings.


\subsection{Token Embedding Interpretability}

Explorations into contextualized embeddings have revealed that a wide range of linguistic and conceptual information is already encoded within individual token vectors. A series of probing studies has demonstrated that syntactic and semantic properties—such as part-of-speech, dependency relations, and even hierarchical sentence structure—can be recovered through relatively simple classifiers \citep{hewitt_structural_2019, conneau_what_2018, voita_information-theoretic_2020, pimentel_information-theoretic_2020}. These findings point to an implicit comprehension of language structure within pretrained models \citep{jawahar_what_2019, tenney_bert_2019}, further supported by analyses of attention mechanisms \citep{clark_what_2019}. Moreover, contextualized embeddings have been found to carry not only grammatical knowledge but also elements of world knowledge and commonsense reasoning \citep{liu2021probing}.


\subsection{Mechanistic Interpretability}

Disentangling dense representation has been a core focus of modern mechanistic interpretability research \citep{bereska_mechanistic_2024}. Probes are limited in terms of what they can explain, as they are aimed at identifying information. \textit{Reconstructive approaches} attempt to decompose the space into individual composition which identifies deeper causal representation. Recent research has aimed to decompose tokens into interpretable components \citep{cunningham2023sparse, bricken2023monosemanticity}. This has typically been done in an unsupervised way, where features were identified post-hoc, through clustering or automated feature analysis \citep{gao2024scaling}. Supervised decomposition approach include a classification loss as a way to align the underlying disentangled representation with a feature \citep{mairal2008supervised, gangeh_supervised_2015}. 

\section{Methods}
\label{sec:framework}

This section presents our theoretical framework. We start by noting that typical sentence embedding models can be broken down into two main stages:  
\begin{enumerate}
    \item \textbf{Token-level processing}: Each input token $t$ in a sentence is mapped to a vector representation $\mathbf{x}_t\in \mathbb{R}^d$. 
    \item \textbf{Pooling.}  Mean pooling then averages the tokens to produce the final embedding $\mathbf{s} = \frac{1}{n}\sum_{t=1}^n \mathbf{x}_t\in\mathbb{R}^d$.
\end{enumerate}
While other pooling strategies exist \citep{xing2025comparativeanalysispoolingmechanisms}, we focus on mean pooling, given its dominant use in industry and research \citep{wang2022text}. Our method, schematized in  Figure~\ref{fig:overview},  will focus on each of these two stages in turn.

\subsection{Token-Level Decomposition}
\label{sec:framework-probe}
A first step towards a mechanistic decomposition would be to extract relevant latent features from tokens. This step implies disentangling representation into interpretable components.

\paragraph{Probes.} Probes are supervised models trained on activations \( \mathbf{h}_i \) to assess whether a target label \( y_i \) (e.g., syntactic role) is recoverable  ~\citep{hewitt_structural_2019, conneau_what_2018, liu2021probing}. Given training set $\{(\mathbf{h}_i, y_i)\}_{1=1}^n$, we optimize our probe \( g_\phi \) with parameters $\phi$ via: 
\[
\min_{\phi} \sum_{i=1}^n \mathcal{L}_{\text{sup}}(g_\phi(\mathbf{h}_i), y_i)
\]
where \( g_\phi \) may be linear (e.g., \( W \mathbf{h}_i + b \)) or non-linear (e.g., a small MLP), and $\cL_{\text{sup}}$ is a choice of supervised loss. This step acts as a diagnostic if key structure is indeed encoded, and if so, if it can be retrieved linearly or not. In most of our cases, $\mathbf{h}_i$ is defined as a token representation $\mathbf{x}_i$ retrieved from an embedding model, and $\mathbf{y_i}$ a token-level annotation (e.g., a part-of-speech tag; in this example, the probe would try to learn to map token-level embeddings onto part-of-speech tags, and we could figure out how well this is possible, both linear or non-linearly).

\paragraph{Supervised Dictionary Learning.}
While probes can reveal whether a particular property is encoded in a representation, they do not provide a full decomposition into independent semantic components \citep{ravichander_probing_2021}. To disentangle the information in each token embedding, we propose supervised dictionary learning: we model each token embedding $\mathbf{x}_t\in\mathbb{R}^d$ as a sparse linear combination of $k$ dictionary atoms from a dictionary $D$, i.e.,
\[
  X \approx DZ \] 
\[   X=[\textbf{x}_1,\ldots, \textbf{x}_T] \in \RR^{d \times T} \] 
\[ D=[\textbf{d}_1, \ldots, \textbf{d}_k] \in\mathbb{R}^{d\times k} \] 
  \[  Z:= [\textbf{z}_1, \ldots, \textbf{z}_T] \in \RR^{k \times T} , \quad \forall i, 
  \|\mathbf{z}_i\|_0 \le \epsilon.
\]
\noi for some small $\epsilon \in \RR_{> 0}$. The sparse code $\mathbf{z}_t$ shows which dictionary atoms are active in a token embedding.
To steer the learned atoms toward interpretable features, we introduce supervision via an auxiliary predictor $f_\theta$ (e.g.\ for syntactic features we have identified with probes). We jointly optimize the dictionary $D$, sparse codes $Z=\{\mathbf{z}_i\}$, and predictor parameters $\theta$ by minimizing: for $i=1,\ldots, T$, 

\[
\begin{aligned}
\min_{D,Z,\theta}\;&\sum_{i=1}^T \bigl\|\mathbf{x}_i - D\,\mathbf{z}_i\bigr\|_2^2 \;+\;\lambda\,\mathcal{L}_{\mathrm{sup}}\bigl(f_\theta(\mathbf{z}_i),y_i\bigr)\\
\text{s.t.}\;&\|\mathbf{z}_i\|_0 \le \epsilon 
\end{aligned}
\]

Here, $\mathcal{L}_{\mathrm{sup}}$ is a supervised loss (e.g.\ cross‐entropy) and $\lambda$ balances reconstruction against predictive accuracy.  This joint objective ensures that the resulting atoms are both reconstructive and predictive of the target labels, hence interpretable \citep{mairal2008supervised, gangeh_supervised_2015}.

\paragraph{Interpretability.}  
This framework delivers interpretability through two complementary mechanisms. First, the supervised dictionary learning objective yields interpretable atoms by nature: each atom’s contribution to the prediction can be directly quantified via the learned predictor weights and supervised loss. Second, for atoms not tied to explicit labels, we apply post‑hoc  techniques, such as activation maximization and concept attribution, to map the features to clearly interpretable elements \citep{bereska_mechanistic_2024}.

\subsection{Pooling Compression}
\label{sec:pooling_compression}

The above supervised dictionary learns decomposed atoms, which encode a set of features, which we guide to reflect specific features.

\paragraph{Mean Pooling} Mean pooling over decomposed representations results in a weighted combination of the dictionary atoms $D$ with the sparse codes $Z$. If a sentence  \textbf{s} contains $T$ tokens $\textbf{x}_t$, \( \mathbf{x}_t \) is decomposed as \( \mathbf{x}_t = \sum_{k=1}^{K} z_{t,k} \mathbf{d}_k \) where \( z_{t,k} \) be the activation of atom \( \mathbf{d}_k \) at position \( t \), then the pooled sentence representation is:
\[ 
\textbf{s}= D\bar{\textbf{z}} = \sum_{k=1}^{K} \bar{z}_k \mathbf{d}_k, \quad \text{and} \quad \bar{z}_k = \frac{1}{T} \sum_{t=1}^{T} z_{t,k}
\]




Atoms that are rarely activated do not meaningfully influence the pooled representation unless their activations are large. An atom can dominate the pooled representation either by being consistently present at moderate levels (high mean, low variance) or by contributing an occasional large spike (moderate mean but high variance). The mean pooling operation does not differentiate these scenarios and sees only the average.
 

\paragraph{Attribution of preserved content.}  

To disentangle these factors and identify interpretable features of the pooled representation \textbf{s}, we assign each atom $d_k$ a contribution to the overall representation:

\[
a_k \;=\;
\underbrace{\bar z_k}_{\text{usage}}
\;\cdot\;
\underbrace{\langle \mathbf{d}_k,\;\mathbf{s}\rangle}_{\text{directional alignment}}
\]

A large positive $a_k$ means that atom $k$ is both frequently activated and points along the overall sentence direction, thereby reinforcing the pooled embedding. The opposite applies for a negative $a_k$. Atoms with $|a_k|$ near zero are either seldom used or orthogonal to the sentence meaning. Normalising $\{a_k\}$ to sum to one lets us rank atoms (or groups such as POS/DEP classes) by their contribution to the overall representation.


\subsection{Research questions}
\label{sec:research_questions}

The above provides a general interpretability framework capable of identifying relevant features both before and after pooling. We now apply the method to address the following research questions:

\begin{itemize}[topsep=0pt,itemsep=2pt,parsep=0pt,partopsep=0pt]
  \item Are linguistic properties linearly present in the token encodings? See \autoref{interpretable_components}
  \item Can dictionary learning isolate these properties into sparse atoms? See \autoref{linear_token}
  \item Which atoms, and which linguistic properties, persist after pooling? See \autoref{pooling_atoms} and  \autoref{pooling_features}.
\end{itemize}

\section{Token-Level Representation Analysis}

We now turn to the empirical decomposition of information encoded in sentence embedding tokens. We run our experiments on the Brown corpus of American English words, which contains a diverse and high-quality corpus with annotations \cite{francis79browncorpus}. We sampled 20000 sentences from the dataset, bringing our corpus to over 140000 tokens. We evaluate three sentence embedding models from HuggingFace: \texttt{multilingual-e5-large} (intfloat), \texttt{all-mpnet-base-v2}, and \texttt{all-MiniLM-L6-v2} (sentence-transformers) \citep{reimers_sentence-bert_2019,  wang2024multilingual}. These span different architectures and training objectives, providing a representative basis for our decomposition analysis. Further experimental details are provided in Appendix~\ref{sec:training_results}.

\subsection{Identifying Interpretable Components}
\label{interpretable_components}

\paragraph{Motivation.} Prior work has demonstrated that sentence representations encode rich syntactic and semantic structure \citep{conneau_what_2018, tenney_bert_2019, jawahar_what_2019}. To understand these representations mechanistically, we aim to identify whether distinct, interpretable components can be reliably isolated, and if such components exist, what methods are best suited to extract them. 

\paragraph{Setup.} To test the above hypothesis, we study the encoding of syntactic information. We tokenize the Brown Corpus  using Stanza \citep{qi2020stanza}, extracting part-of-speech and dependency labels. When words are split into subwords, we aggregate them into a single token representation following the approach of \citet{acs_subword_2021}. We train linear and non-linear probes (formal definition in \S \ref{sec:framework-probe})  to predict part-of-speech, dependency tags and position of token in the sentence from token-level embeddings. We then identify structure by running SVD on the weights of this probe, as a means of dimensionality reduction, to determine whether there are signs of structure at this level. As two baselines, we run the predictions on i) shuffled inputs and also perform ii) random prediction.

\begin{figure}[t]
    \centering
    \includegraphics[width=\columnwidth]{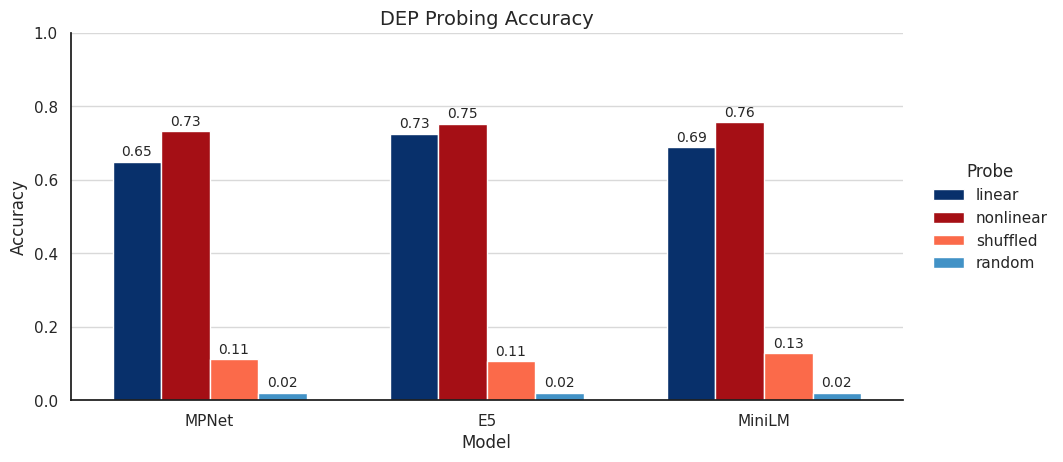}
    \caption{\small Dependency probing accuracy of linear, nonlinear, shuffled, and random probes across three embedding models (MPNet, E5, MiniLM). Shuffled and random baselines remain near chance.
} 
    \label{fig:dep_probe}
\end{figure}

\begin{figure}[t]
    \centering
    \includegraphics[width=\columnwidth]{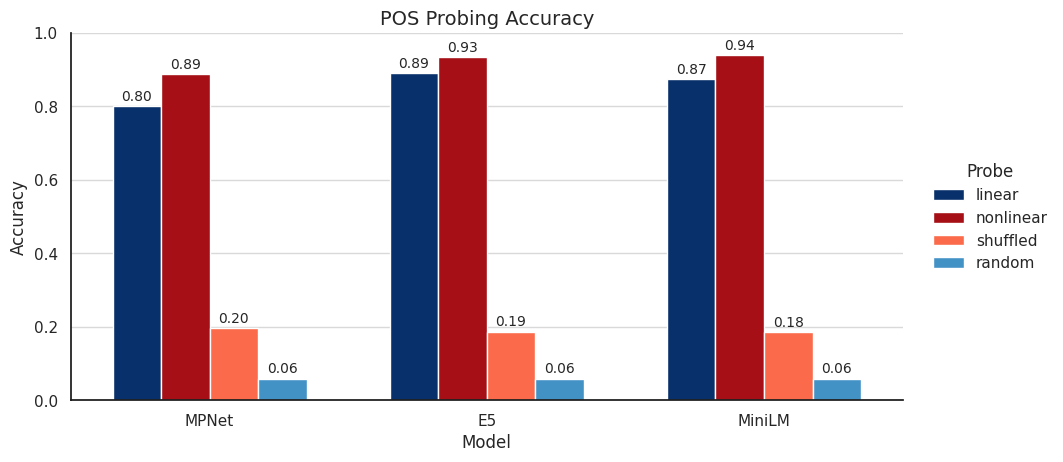}
    \caption{\small Part‑of‑speech probing accuracy of linear, nonlinear, shuffled, and random probes across our three embedding models (multilingual-e5-large,
all-mpnet-base-v2, all-MiniLM-L6-v2). Nonlinear probes outperforms by a slight amount all other settings, with shuffled and random probes showing minimal signal.
} 
    \label{fig:pos_probe}
\end{figure}

\begin{figure}[t]
    \centering
    \includegraphics[width=\columnwidth]{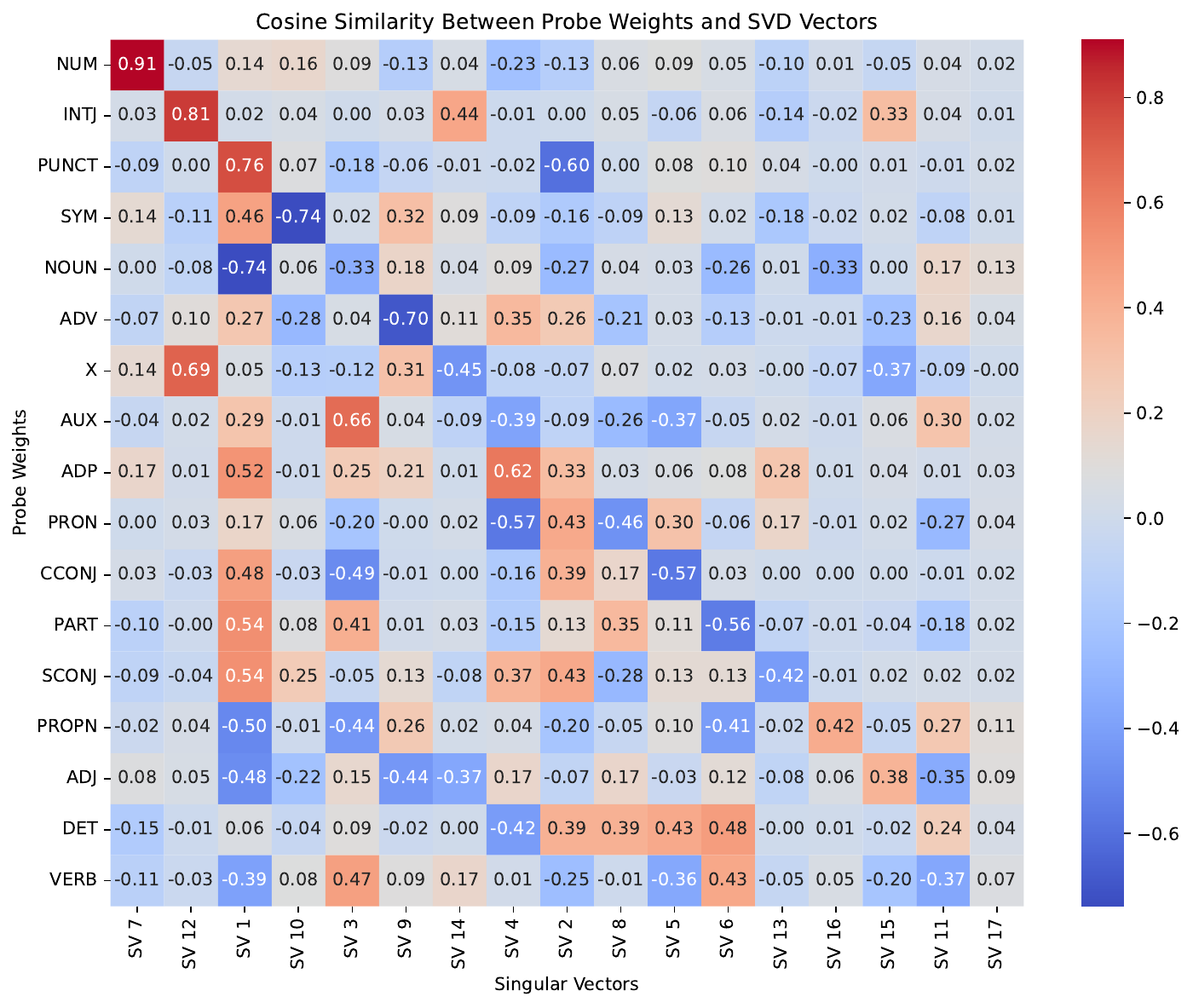}
    \caption{\small Heatmap showing cosine similarity between probe weights and SVD basis vectors for all-MiniLM-L6-v2. We distinguish semantic subspaces, or clusters of activations across probes.
} 
    \label{fig:svd_probe_heatmap}
\end{figure}

\paragraph{Results.} We observe that both linear and nonlinear probes attain high accuracy across all models for core syntactic features. For POS tagging, linear probes achieve up to 0.89 (MiniLM), closely matched by nonlinear probes at 0.94, with similarly tight margins across MPNet and E5 models (Figure \ref{fig:pos_probe}). This suggests that most syntactic information is linearly encoded within the sentence embeddings. In contrast, for dependency labeling (DEP), the nonlinear probe maintains a modest lead, (Figure \ref{fig:dep_probe}), indicating mild non-linear entanglement for specific features. We also find interpretable internal structure in the linear probes. For instance, the 12th singular vector aligns strongly with the POS class INTJ, a relatively rare tag, suggesting that less dominant components specialize in capturing specific, often sparsely encoded properties. All of the values for SVD similarities are shown in Figure \ref{fig:svd_probe_heatmap}.

\paragraph{Takeaway.} The studied sentence embedding models largely encode key syntactic information linearly, as non-linear probes are only slightly more accurate. The structure of the SVD space is a strong indication of the possibility of decomposition in interpretable subspaces. 

\subsection{Supervised Dictionary Learning}
\label{linear_token}
\paragraph{Motivation.} Probes have shown us what type of information was encoded and how it was best retrieved. We now aim to decompose token-level representations into latent features. If successful, this would allow us to identify interpretable atoms that encode interpretable features, including those we identified with probes. Our goal is to determine whether token representations can be reliably reconstructed from such a basis.

\paragraph{Setup.} We apply supervised dictionary learning to our dataset in order to decompose token representations into interpretable components. Each contextualized embedding is reconstructed from a learned dictionary of semantic atoms, while also being used to predict linguistic labels such as part-of-speech and dependency relations. To further constrain the atoms, we align the reconstructions with static (pre-contextualized) embeddings from the same transformer, allowing us to disentangle the static word representation from contextual variation. Our objective combines four terms: (1) reconstruction loss over contextual embeddings, (2) cross-entropy loss for POS and dependency prediction, (3) reconstruction loss over a static embedding (the pre-contextualized word vector), and (4) a sparsity penalty on the atom activations. 

\begin{figure}[t]
    \centering
    \includegraphics[width=\columnwidth]{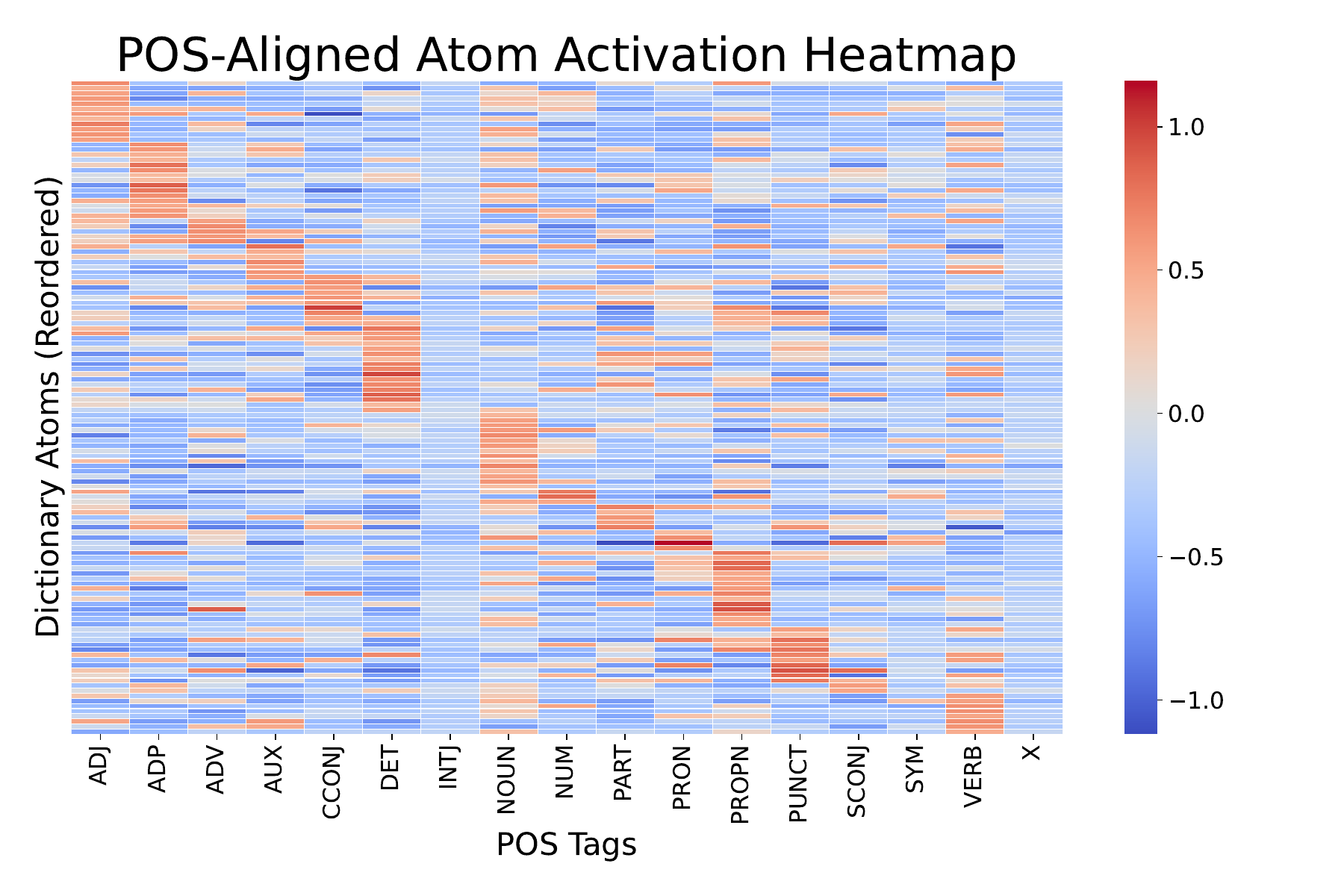}
    \caption{\small Heatmap of POS-specific activation patterns across dictionary atoms for the best-performing hyperparameter configuration of all-MiniLM-L6-v2 with a dictionary size $k$ of 64, and a linear encoder. }
    
    \label{fig:atom_heatmap}
\end{figure}

\begin{figure}[t]
    \centering
    \includegraphics[width=\columnwidth]{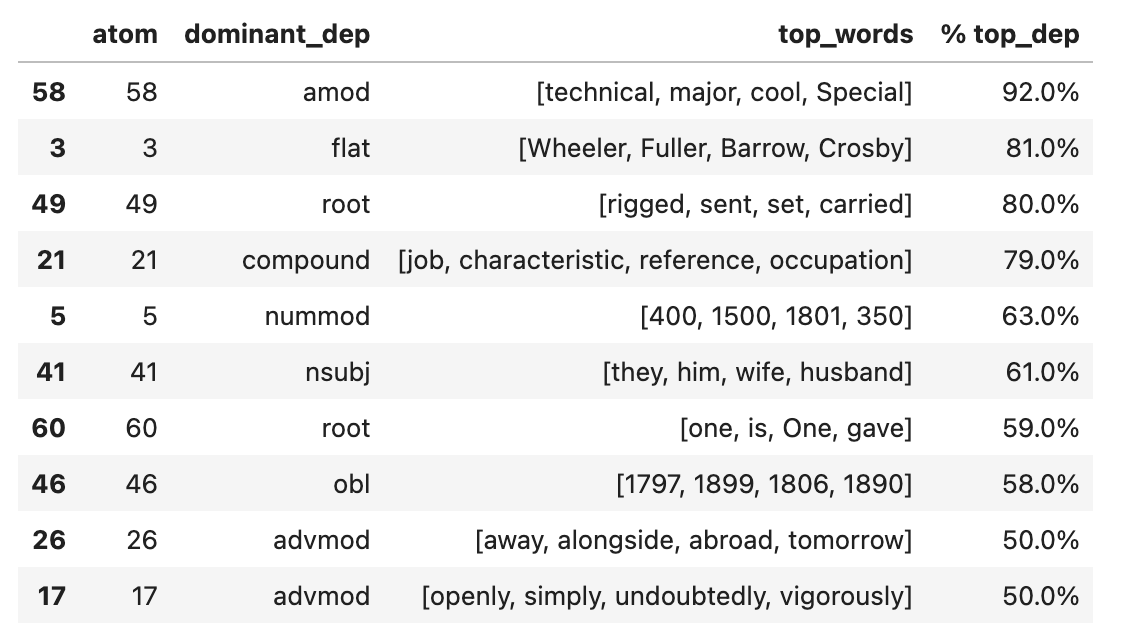}
    \caption{\small Top atoms for our best performing dictionary configuration for all-MiniLM-L6-v2. Our token level atoms are interpretable, and specialize in specific syntactic features. 
    The ten atoms with the highest label‑assignment confidence (up to 92\%) each specialize in a single dependency relation—e.g. atom 58 detects adjectival modifiers. 
} 
    \label{fig:dominant_pos}
\end{figure}

\paragraph{Results.}
We evaluate multiple dictionary configurations, both linear and nonlinear. As detailed in the Appendix~\ref{sec:training_results}, linear dictionaries perform comparably to their nonlinear counterparts on our task, suggesting that much of the structure can already be captured in a linear subspace. Interestingly, meaningful representations are preserved even when compressing to 64 dimensions, indicating that the model can retain salient features in a compact form. Our objective was to assess if interpretable features could be detected. Even at moderate sparsity levels, we identify interpretable structures in the learned representations of linear dictionary decompositions.  Figure~\ref{fig:atom_heatmap} illustrates this by showing the activation patterns of selected atoms across POS tags, highlighting the dictionary’s capacity to encode syntactic information, and reconstruct the activations. These atoms encode stable semantic structures (e.g., adjectives for atom 58, numerals for atom 5) as seen in Figure~\ref{fig:dominant_pos}. 

\paragraph{Takeaway.}
Dictionary atoms effectively encode structured information and enable interpretable subspace decomposition. Unlike probing, we identified the specific regions in representation space where this information is encoded, going beyond classical techniques \citep{conneau_what_2018, tenney_bert_2019, belinkov2021probingclassifierspromisesshortcomings, opitz_interpretable_2025}.



\section{Information Retention: Pooling Effects}
\label{sec:effect_pooling}
In order to understand the final sentence representation, we have to assess how mean pooling compresses token information into a single representation. Meal pooling compresses a sequence of token codes $\crbr{\textbf{z}_t}$ into the sentence code $\bar{\textbf{z}}$. We ask two questions: \textbf{(1) Atom level}: which dictionary atoms $\textbf{d}_k$ survive this averaging? \textbf{(2) Feature level}: which linguistic categories (e.g. POS, DEP) contribute most to the final embedding ? 

For clarity, we will run our experiments on the best performing configuration of  all-MiniLM-L6-v2, with a dictionary size of 64 and a linear encoder. 

    

\subsection{Pooling Across Atoms}
\label{pooling_atoms}

\paragraph{Motivation.} The above section has shown that we have interpretable atoms before pooling. We investigate how individual dictionary atoms contribute to the pooled vector: i.e. which atoms persist, which are suppressed. Mean pooling depends on the directional alignment of the atoms with the mean, complemented by the usage of the atoms. 

\paragraph{Setup.}  We compute the mean and variance across all the tokens. For each dictionary atom \( k \), we track its mean activation and variance across the sentence. We then assign each atom a contribution, following \autoref{sec:pooling_compression}. These metrics allow us to quantify the roles of specific atoms on the final representation. We then map these to interpretable features. 

\begin{figure}[t]
    \centering
    \begin{subfigure}{\columnwidth}
        \centering
        \includegraphics[width=\linewidth]{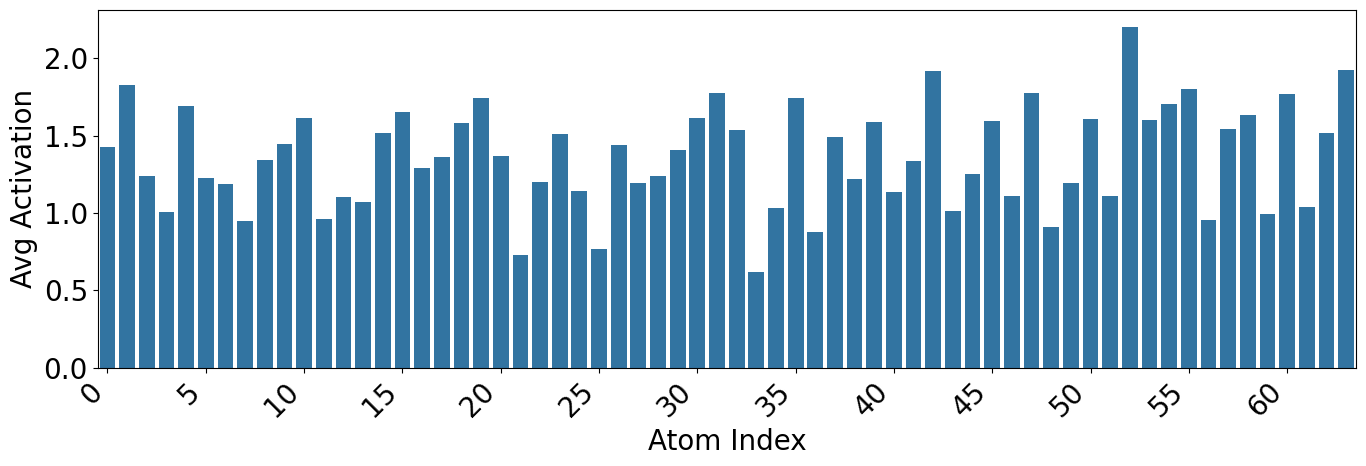}
        \caption{\small Top atom activations for the best-performing hyperparameter configuration of all-MiniLM-L6-v2. Most atoms maintain baseline activations across tokens, but a small subset (e.g. atoms 52) exhibit substantially elevated mean absolute activations. 
 }
     \label{fig:mean_graph}
    \end{subfigure}

    \begin{subfigure}{\columnwidth}
        \centering
        \includegraphics[width=\linewidth]{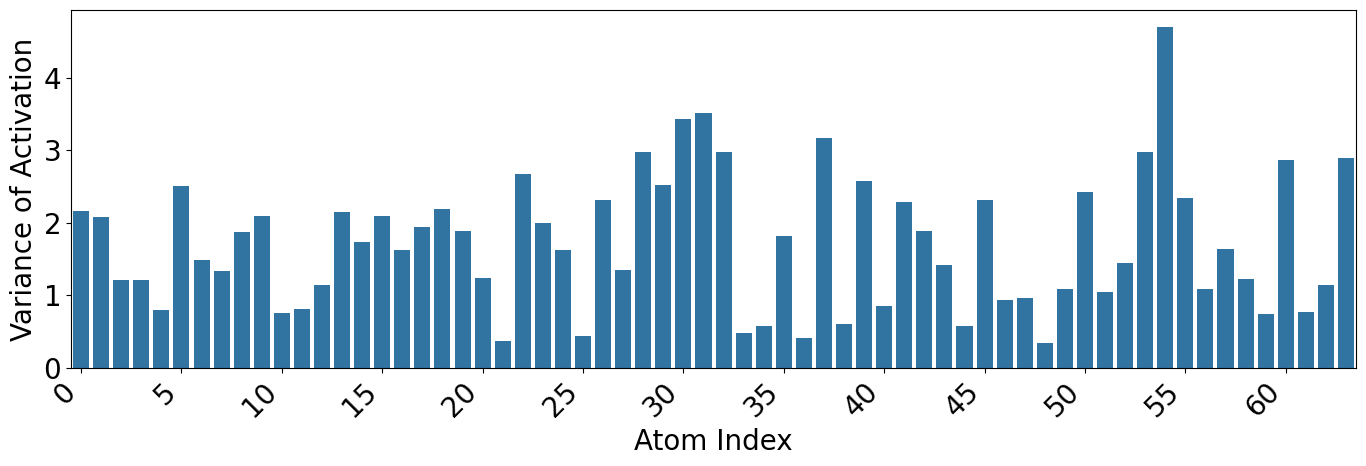}
        \caption{\small The variance across atoms differs widely, which highlights specific features which vary more widely across sentences}
     \label{fig:variance_graph}
    \end{subfigure}
    \caption{Activation mean and variance across dictionary atoms for the best-performing hyperparameter configuration of all-MiniLM-L6-v2}
\end{figure}

\begin{figure}[t]
    \centering
    \includegraphics[width=\columnwidth]{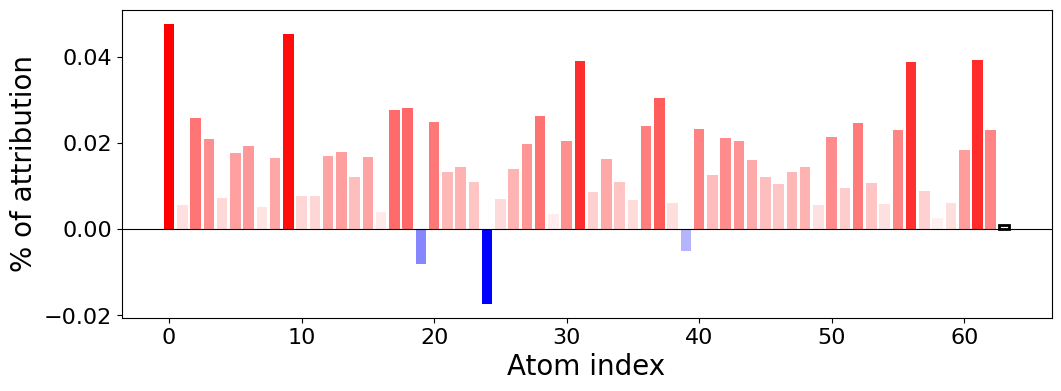}
    \caption{\small Average contributions of atoms on our full dataset for our best trial of all-MiniLM-L6-v2. A negative contribution implies that the atom cancels others, which can be explained by overlap between features, as the dictionary is not fully orthogonal. We highlighted the last atom for visibility. 
} 
    \label{fig:contribution_dataset}
\end{figure}

\paragraph{Results.}
Atoms in our dictionary exhibit relatively stable mean activation values but show notable differences in variance. This suggests that while some atoms are broadly active across many tokens, others are more selective, activating only for specific lexical or syntactic contexts. As illustrated in Figures~\ref{fig:mean_graph} and~\ref{fig:variance_graph}, this pattern reflects a spectrum of generality and specificity among atoms. The distribution of atom-level contributions is not uniform: although many atoms contribute moderately, a subset (such as atoms 0, 9, and 62; see \autoref{fig:contribution_dataset}) clearly dominates the final sentence representation. These contributions are interpretable. For example, atom 5, associated with numerical tokens, has limited influence. Similarly, atom 58, which activates for adverbs describing actions, contributes weakly, as seen in \autoref{fig:dominant_pos} and \autoref{fig:contribution_dataset}.

\paragraph{Takeaway.} 
We were able to identify the precise contributions of atoms and map them to interpretable features. Our framework provides a means to show which features contribute to an embedding in impactful ways, and which don't. 


\subsection{Interpreting Pooled Features}
\label{pooling_features}

\paragraph{Motivation.} In the subsection above, we have identified the contribution of individual atoms to the pooled sentence representation. We now focus on identifying the contribution of specific features. As per \autoref{sec:research_questions}, we will use the features we previously detected (POS, DEP) to assess if our method can also help us identify chosen features. 

\begin{figure}[t]
    \centering
    \includegraphics[width=\columnwidth]{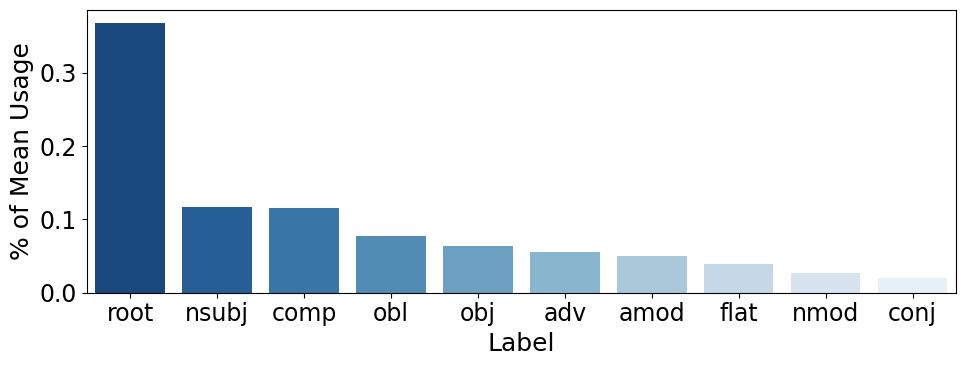}
    \caption{\small Relative contributions of the top 12 dependencies tag to the overall sentence representation on our best trial of all-MiniLM-L6-v2.
} 
    \label{fig:contribution_dep}
\end{figure}

\begin{figure}[t]
    \centering
    \includegraphics[width=\columnwidth]{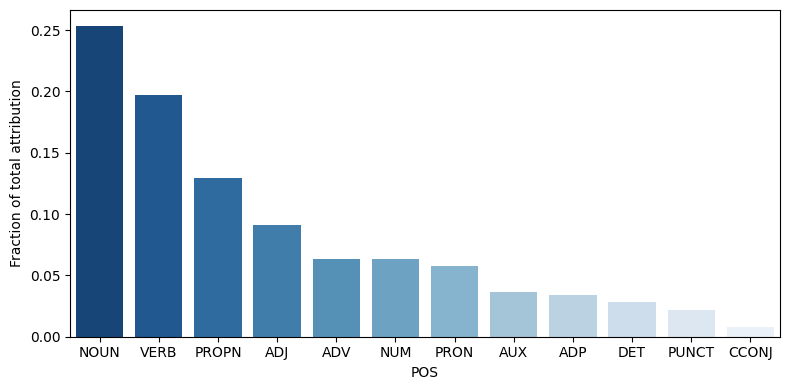}
    \caption{\small Relative contributions of part-of-speech tags to the overall sentence representation in our best-performing run using all-MiniLM-L6-v2. The plot includes only tags that account for more than 2\% of the total attribution.}

\label{fig:contribution_pos}
\end{figure}

\paragraph{Setup.} 
Following \autoref{sec:pooling_compression}, we compute a normalised attribution vector for each sentence, indicating the contribution of each atom to the pooled embedding. Aggregating these vectors across the dataset yields a global importance score for each atom. To account for the fact that atoms may respond to multiple syntactic roles, we assign each atom a set of fractional weights~$\pi_{i,c}$, reflecting how often it is activated by tokens of class~$c$ (POS or DEP). We then compute the total contribution of each class by summing the corresponding fractions of atom-level contributions.

\paragraph{Results.}
We find that the most influential parts of speech are nouns, verbs, and proper nouns, which aligns with their central role as semantically rich elements in sentence meaning (\autoref{fig:contribution_pos}). In contrast, categories such as punctuation contribute far less to the final representation. These contributions mirror semantic importance, and we observe a long tail of POS tags that each account for less than 5\% of the total attribution. A similar pattern emerges for dependency labels \autoref{fig:contribution_dep}, where root, subjects and objects dominate attribution.

\paragraph{Takeaway.}
Our method recovers patterns consistent with the literature (\autoref{sec:background}), while assigning precise attribution scores to features. It captures, for example, the dominant influence of central categories like nouns and verbs. In our above example with all-MiniLM-L6-v2, we are able to attribute features with a linear dictionary. 

From this experiment we can also draw a linguistic take-away: While the strong saliency of POS tags like Nouns, Verbs seems natural, the strong saliency of the grammatical \textit{root} relation, a linguistically motivated but purely theoretic entity, empirically seems to support both dependency grammar design choices \citep{tesniere1959elements} and also theories that ``core-semantics'' are closest to the root of a sentence \citep[c.f.][]{cai-lam-2019-core}. 

\section{Conclusion}

Our work lays out a proof of concept for identifying interpretable features in sentence representations, proposing the first mechanistic-interpretablilty based approach. Through experiments across different model types, we show that it is possible to recover fine-grained features and trace how they are (often \textit{linearly}) preserved merged into a pooled sentence representation.

This contributes to bridging recent advances in mechanistic interpretability with sentence-level embedding analysis, allowing for a more granular understanding of the representations. Sentence embeddings encode structured knowledge, and our method enables a scalable extraction of these features. This, in turn, makes it feasible to develop a mechanistic account of the information embedded in these representations, as well as to understand the role that pooling strategies play in preserving or diluting this information. Such insights can enrich the broader field of interpretable sentence embeddings and understanding model-based semantic representation of linguistic phenomena \citep{opitz_interpretable_2025, bereska_mechanistic_2024}.

As such, we believe that our approach holds particular promise not only for downstream tasks such as information retrieval, sentence similarity, and retrieval-augmented generation, but also certain kinds of linguistic investigations, e.g., towards saliency of linguistic units for semantics, or studies of composition and construction. Indeed, interpretability of NLP models is not only useful for mitigating bias and increasing model transparency, but also for improving performance and, as we have seen, it may even foster linguistic studies and understanding \citep{lewis2020retrieval, hambarde2023information, gao2023retrieval}.

\section*{Limitations}

This work is a bluerprint and as such we highlight several important limitations and possible future workstreams:
\begin{enumerate}
    \item \textbf{Model coverage.} While we tested our approach on three widely-used sentence-embedding models, a broader evaluation would strengthen the evidence base. 
    Applying the same diagnostic suite to different models and architectures (e.g. GTE \citep{zhang-etal-2024-mgte}), would yield further learnings.
    \item \textbf{Corpus scale.} Our 140 k-token Brown-corpus sample suffices for a proof-of-concept, but scaling to larger and domain-diverse corpora is an important endeavor for future work. It would also allow to explore linguistic phenomena as represented by models, at scale, including cross-lingual and multi-lingual ones.
    \item  \textbf{Pooling variants}: We analysed mean pooling because it remains the de-facto default in research and industry. Extending our analysis to max pooling or attention pooling can quantify how each mechanism re-weights sparsity patterns. It is also important for developing a framework to give a better notion of dominant atoms.
\item \textbf{Token position.} Preliminary plots suggest that early tokens fire fewer high-information atoms than later ones. We suspect that this is because later tokens are more contextualized and encode a richer representation of the sentence as a whole. More position-aware diagnostics and experiments could help clarify how pooling equalises (or fails to equalise) this asymmetry
\end{enumerate}




\bibliography{acl_latex}

\begin{thebibliography}{49}
\providecommand{\natexlab}[1]{#1}

\bibitem[{Akiba et~al.(2019)Akiba, Sano, Yanase, Ohta, and Koyama}]{optuna_2019}
Takuya Akiba, Shotaro Sano, Toshihiko Yanase, Takeru Ohta, and Masanori Koyama. 2019.
\newblock Optuna: A next-generation hyperparameter optimization framework.
\newblock In \emph{Proceedings of the 25th {ACM} {SIGKDD} International Conference on Knowledge Discovery and Data Mining}.

\bibitem[{Arora et~al.(2017)Arora, Liang, and Ma}]{arora2017simple}
Sanjeev Arora, Yingyu Liang, and Tengyu Ma. 2017.
\newblock A simple but tough-to-beat baseline for sentence embeddings.
\newblock In \emph{International conference on learning representations}.

\bibitem[{Barnab{\`o} et~al.(2023)Barnab{\`o}, Uva, Pollastrini, Rubagotti, and Bernardi}]{barnabo-etal-2023-supervised}
Giorgio Barnab{\`o}, Antonio Uva, Sandro Pollastrini, Chiara Rubagotti, and Davide Bernardi. 2023.
\newblock \href {https://doi.org/10.18653/v1/2023.findings-ijcnlp.36} {Supervised clustering loss for clustering-friendly sentence embeddings: an application to intent clustering}.
\newblock In \emph{Findings of the Association for Computational Linguistics: IJCNLP-AACL 2023 (Findings)}, pages 412--430, Nusa Dua, Bali. Association for Computational Linguistics.

\bibitem[{Belinkov(2021)}]{belinkov2021probingclassifierspromisesshortcomings}
Yonatan Belinkov. 2021.
\newblock \href {https://arxiv.org/abs/2102.12452} {Probing classifiers: Promises, shortcomings, and advances}.
\newblock \emph{Preprint}, arXiv:2102.12452.

\bibitem[{Benara et~al.(2024)Benara, Singh, Morris, Antonello, Stoica, Huth, and Gao}]{benara2024crafting}
Vinamra Benara, Chandan Singh, John~X Morris, Richard~J Antonello, Ion Stoica, Alexander~G Huth, and Jianfeng Gao. 2024.
\newblock Crafting interpretable embeddings for language neuroscience by asking llms questions.
\newblock \emph{Advances in neural information processing systems}, 37:124137.

\bibitem[{Bereska and Gavves(2024)}]{bereska_mechanistic_2024}
Leonard Bereska and Efstratios Gavves. 2024.
\newblock \href {https://doi.org/10.48550/arXiv.2404.14082} {Mechanistic {Interpretability} for {AI} {Safety} -- {A} {Review}}.
\newblock ArXiv:2404.14082 [cs].

\bibitem[{Bommasani et~al.(2020)Bommasani, Davis, and Cardie}]{bommasani-etal-2020-interpreting}
Rishi Bommasani, Kelly Davis, and Claire Cardie. 2020.
\newblock \href {https://doi.org/10.18653/v1/2020.acl-main.431} {{I}nterpreting {P}retrained {C}ontextualized {R}epresentations via {R}eductions to {S}tatic {E}mbeddings}.
\newblock In \emph{Proceedings of the 58th Annual Meeting of the Association for Computational Linguistics}, pages 4758--4781, Online. Association for Computational Linguistics.

\bibitem[{Bricken et~al.(2023)Bricken, Templeton, Batson, Chen, Jermyn, Conerly, Turner, Anil, Denison, Askell, Lasenby, Wu, Kravec, Schiefer, Maxwell, Joseph, Hatfield-Dodds, Tamkin, Nguyen, McLean, Burke, Hume, Carter, Henighan, and Olah}]{bricken2023monosemanticity}
Trenton Bricken, Adly Templeton, Joshua Batson, Brian Chen, Adam Jermyn, Tom Conerly, Nick Turner, Cem Anil, Carson Denison, Amanda Askell, Robert Lasenby, Yifan Wu, Shauna Kravec, Nicholas Schiefer, Tim Maxwell, Nicholas Joseph, Zac Hatfield-Dodds, Alex Tamkin, Karina Nguyen, Brayden McLean, Josiah~E Burke, Tristan Hume, Shan Carter, Tom Henighan, and Christopher Olah. 2023.
\newblock Towards monosemanticity: Decomposing language models with dictionary learning.
\newblock \emph{Transformer Circuits Thread}.
\newblock Https://transformer-circuits.pub/2023/monosemantic-features/index.html.

\bibitem[{Cai and Lam(2019)}]{cai-lam-2019-core}
Deng Cai and Wai Lam. 2019.
\newblock \href {https://doi.org/10.18653/v1/D19-1393} {Core semantic first: A top-down approach for {AMR} parsing}.
\newblock In \emph{Proceedings of the 2019 Conference on Empirical Methods in Natural Language Processing and the 9th International Joint Conference on Natural Language Processing (EMNLP-IJCNLP)}, pages 3799--3809, Hong Kong, China. Association for Computational Linguistics.

\bibitem[{Clark et~al.(2019)Clark, Khandelwal, Levy, and Manning}]{clark_what_2019}
Kevin Clark, Urvashi Khandelwal, Omer Levy, and Christopher~D. Manning. 2019.
\newblock \href {https://doi.org/10.48550/arXiv.1906.04341} {What {Does} {BERT} {Look} {At}? {An} {Analysis} of {BERT}'s {Attention}}.
\newblock ArXiv:1906.04341 [cs].

\bibitem[{Conneau et~al.(2018)Conneau, Kruszewski, Lample, Barrault, and Baroni}]{conneau_what_2018}
Alexis Conneau, German Kruszewski, Guillaume Lample, Loïc Barrault, and Marco Baroni. 2018.
\newblock \href {https://doi.org/10.18653/v1/P18-1198} {What you can cram into a single \$\&!\#* vector: {Probing} sentence embeddings for linguistic properties}.
\newblock In \emph{Proceedings of the 56th {Annual} {Meeting} of the {Association} for {Computational} {Linguistics} ({Volume} 1: {Long} {Papers})}, pages 2126--2136, Melbourne, Australia. Association for Computational Linguistics.

\bibitem[{Cunningham et~al.(2023)Cunningham, Ewart, Riggs, Huben, and Sharkey}]{cunningham2023sparse}
Hoagy Cunningham, Aidan Ewart, Logan Riggs, Robert Huben, and Lee Sharkey. 2023.
\newblock Sparse autoencoders find highly interpretable features in language models.
\newblock \emph{arXiv:2309.08600}.

\bibitem[{Fodor et~al.(2025)Fodor, Deyne, and Suzuki}]{fodorchallenge}
James Fodor, Simon~De Deyne, and Shinsuke Suzuki. 2025.
\newblock \href {https://doi.org/10.1162/coli_a_00536} {Compositionality and sentence meaning: Comparing semantic parsing and transformers on a challenging sentence similarity dataset}.
\newblock \emph{Computational Linguistics}, 51(1):139--190.

\bibitem[{Francis and Kucera(1979)}]{francis79browncorpus}
W.~N. Francis and H.~Kucera. 1979.
\newblock \href {http://icame.uib.no/brown/bcm.html} {Brown corpus manual}.
\newblock Technical report, Department of Linguistics, Brown University, Providence, Rhode Island, US.

\bibitem[{Gangeh et~al.(2015)Gangeh, Farahat, Ghodsi, and Kamel}]{gangeh_supervised_2015}
Mehrdad~J. Gangeh, Ahmed~K. Farahat, Ali Ghodsi, and Mohamed~S. Kamel. 2015.
\newblock \href {https://doi.org/10.48550/arXiv.1502.05928} {Supervised {Dictionary} {Learning} and {Sparse} {Representation}-{A} {Review}}.
\newblock ArXiv:1502.05928.

\bibitem[{Gao et~al.(2024)Gao, la~Tour, Tillman, Goh, Troll, Radford, Sutskever, Leike, and Wu}]{gao2024scaling}
Leo Gao, Tom~Dupr{\'e} la~Tour, Henk Tillman, Gabriel Goh, Rajan Troll, Alec Radford, Ilya Sutskever, Jan Leike, and Jeffrey Wu. 2024.
\newblock Scaling and evaluating sparse autoencoders.
\newblock \emph{arXiv:2406.04093}.

\bibitem[{Gao et~al.(2021)Gao, Yao, and Chen}]{gao2021simcse}
Tianyu Gao, Xingcheng Yao, and Danqi Chen. 2021.
\newblock Simcse: Simple contrastive learning of sentence embeddings.
\newblock \emph{arXiv:2104.08821}.

\bibitem[{Gao et~al.(2023)Gao, Xiong, Gao, Jia, Pan, Bi, Dai, Sun, and Wang}]{gao2023retrieval}
Yunfan Gao, Yun Xiong, Xinyu Gao, Kangxiang Jia, Jinliu Pan, Yuxi Bi, Yi~Dai, Jiawei Sun, and Haofen Wang. 2023.
\newblock Retrieval-augmented generation for large language models: A survey.
\newblock \emph{arXiv preprint arXiv:2312.10997}.

\bibitem[{Hambarde and Proenca(2023)}]{hambarde2023information}
Kailash~A Hambarde and Hugo Proenca. 2023.
\newblock Information retrieval: recent advances and beyond.
\newblock \emph{IEEE Access}.

\bibitem[{Han et~al.(2023)Han, Liu, and Wang}]{han2023comprehensive}
Yikun Han, Chunjiang Liu, and Pengfei Wang. 2023.
\newblock A comprehensive survey on vector database: Storage and retrieval technique, challenge.
\newblock \emph{arXiv:2310.11703}.

\bibitem[{Hewitt and Manning(2019)}]{hewitt_structural_2019}
John Hewitt and Christopher~D. Manning. 2019.
\newblock \href {https://doi.org/10.18653/v1/N19-1419} {A {Structural} {Probe} for {Finding} {Syntax} in {Word} {Representations}}.
\newblock In \emph{Proceedings of the 2019 {Conference} of the {North} {American} {Chapter} of the {Association} for {Computational} {Linguistics}: {Human} {Language} {Technologies}, {Volume} 1 ({Long} and {Short} {Papers})}, pages 4129--4138, Minneapolis, Minnesota. Association for Computational Linguistics.

\bibitem[{Huang et~al.(2023)Huang, Yao, Song, Zhang, Chen, and Yu}]{huang_bridging_2023}
James~Y. Huang, Wenlin Yao, Kaiqiang Song, Hongming Zhang, Muhao Chen, and Dong Yu. 2023.
\newblock \href {https://doi.org/10.48550/arXiv.2305.14599} {Bridging {Continuous} and {Discrete} {Spaces}: {Interpretable} {Sentence} {Representation} {Learning} via {Compositional} {Operations}}.
\newblock ArXiv:2305.14599 [cs].

\bibitem[{Jawahar et~al.(2019)Jawahar, Sagot, and Seddah}]{jawahar_what_2019}
Ganesh Jawahar, Benoît Sagot, and Djamé Seddah. 2019.
\newblock \href {https://doi.org/10.18653/v1/P19-1356} {What {Does} {BERT} {Learn} about the {Structure} of {Language}?}
\newblock In \emph{Proceedings of the 57th {Annual} {Meeting} of the {Association} for {Computational} {Linguistics}}, pages 3651--3657, Florence, Italy. Association for Computational Linguistics.

\bibitem[{Kashyap et~al.(2023)Kashyap, Nguyen, Schlegel, Winkler, Ng, and Poria}]{kashyap2023comprehensive}
Abhinav~Ramesh Kashyap, Thanh-Tung Nguyen, Viktor Schlegel, Stefan Winkler, See-Kiong Ng, and Soujanya Poria. 2023.
\newblock A comprehensive survey of sentence representations: From the bert epoch to the chatgpt era and beyond.
\newblock \emph{arXiv:2305.12641}.

\bibitem[{Lewis et~al.(2020)Lewis, Perez, Piktus, Petroni, Karpukhin, Goyal, K{\"u}ttler, Lewis, Yih, Rockt{\"a}schel et~al.}]{lewis2020retrieval}
Patrick Lewis, Ethan Perez, Aleksandra Piktus, Fabio Petroni, Vladimir Karpukhin, Naman Goyal, Heinrich K{\"u}ttler, Mike Lewis, Wen-tau Yih, Tim Rockt{\"a}schel, et~al. 2020.
\newblock Retrieval-augmented generation for knowledge-intensive nlp tasks.
\newblock \emph{Advances in Neural Information Processing Systems}, 33:9459--9474.

\bibitem[{Li et~al.(2020)Li, Zhou, He, Wang, Yang, and Li}]{li2020sentence}
Bohan Li, Hao Zhou, Junxian He, Mingxuan Wang, Yiming Yang, and Lei Li. 2020.
\newblock On the sentence embeddings from pre-trained language models.
\newblock \emph{arXiv:2011.05864}.

\bibitem[{Liu et~al.(2021)Liu, Wang, Kasai, Hajishirzi, and Smith}]{liu2021probing}
Leo~Z Liu, Yizhong Wang, Jungo Kasai, Hannaneh Hajishirzi, and Noah~A Smith. 2021.
\newblock Probing across time: What does roberta know and when?
\newblock \emph{arXiv:2104.07885}.

\bibitem[{Liu et~al.(2020)Liu, Kusner, and Blunsom}]{liu2020survey}
Qi~Liu, Matt~J Kusner, and Phil Blunsom. 2020.
\newblock A survey on contextual embeddings.
\newblock \emph{arXiv:2003.07278}.

\bibitem[{Mairal et~al.(2008)Mairal, Ponce, Sapiro, Zisserman, and Bach}]{mairal2008supervised}
Julien Mairal, Jean Ponce, Guillermo Sapiro, Andrew Zisserman, and Francis Bach. 2008.
\newblock Supervised dictionary learning.
\newblock \emph{Advances in neural information processing systems}, 21.

\bibitem[{Moeller et~al.(2023)Moeller, Nikolaev, and Pad{\'o}}]{moeller-etal-2023-attribution}
Lucas Moeller, Dmitry Nikolaev, and Sebastian Pad{\'o}. 2023.
\newblock \href {https://doi.org/10.18653/v1/2023.emnlp-main.980} {An attribution method for {S}iamese encoders}.
\newblock In \emph{Proceedings of the 2023 Conference on Empirical Methods in Natural Language Processing}, pages 15818--15827, Singapore. Association for Computational Linguistics.

\bibitem[{Moeller et~al.(2024)Moeller, Nikolaev, and Pad{\'o}}]{moeller-etal-2024-approximate}
Lucas Moeller, Dmitry Nikolaev, and Sebastian Pad{\'o}. 2024.
\newblock \href {https://aclanthology.org/2024.eacl-long.125} {Approximate attributions for off-the-shelf {S}iamese transformers}.
\newblock In \emph{Proceedings of the 18th Conference of the European Chapter of the Association for Computational Linguistics (Volume 1: Long Papers)}, pages 2059--2071, St. Julian{'}s, Malta. Association for Computational Linguistics.

\bibitem[{Nastase and Merlo(2024)}]{nastase-merlo-2024-tracking}
Vivi Nastase and Paola Merlo. 2024.
\newblock \href {https://aclanthology.org/2024.repl4nlp-1.15} {Tracking linguistic information in transformer-based sentence embeddings through targeted sparsification}.
\newblock In \emph{Proceedings of the 9th Workshop on Representation Learning for NLP (RepL4NLP-2024)}, pages 203--214, Bangkok, Thailand. Association for Computational Linguistics.

\bibitem[{Opitz and Frank(2022)}]{opitz_sbert_2022}
Juri Opitz and Anette Frank. 2022.
\newblock \href {https://doi.org/10.18653/v1/2022.aacl-main.48} {{SBERT} studies {Meaning} {Representations}: {Decomposing} {Sentence} {Embeddings} into {Explainable} {Semantic} {Features}}.
\newblock In \emph{Proceedings of the 2nd {Conference} of the {Asia}-{Pacific} {Chapter} of the {Association} for {Computational} {Linguistics} and the 12th {International} {Joint} {Conference} on {Natural} {Language} {Processing} ({Volume} 1: {Long} {Papers})}, pages 625--638, Online only. Association for Computational Linguistics.

\bibitem[{Opitz et~al.(2025)Opitz, Möller, Michail, and Clematide}]{opitz_interpretable_2025}
Juri Opitz, Lucas Möller, Andrianos Michail, and Simon Clematide. 2025.
\newblock \href {https://doi.org/10.48550/arXiv.2502.14862} {Interpretable {Text} {Embeddings} and {Text} {Similarity} {Explanation}: {A} {Primer}}.
\newblock ArXiv:2502.14862 [cs].

\bibitem[{Pimentel et~al.(2020)Pimentel, Valvoda, Maudslay, Zmigrod, Williams, and Cotterell}]{pimentel_information-theoretic_2020}
Tiago Pimentel, Josef Valvoda, Rowan~Hall Maudslay, Ran Zmigrod, Adina Williams, and Ryan Cotterell. 2020.
\newblock \href {https://doi.org/10.18653/v1/2020.acl-main.420} {Information-{Theoretic} {Probing} for {Linguistic} {Structure}}.
\newblock In \emph{Proceedings of the 58th {Annual} {Meeting} of the {Association} for {Computational} {Linguistics}}, pages 4609--4622, Online. Association for Computational Linguistics.

\bibitem[{Qi et~al.(2020)Qi, Zhang, Zhang, Bolton, and Manning}]{qi2020stanza}
Peng Qi, Yuhao Zhang, Yuhui Zhang, Jason Bolton, and Christopher~D Manning. 2020.
\newblock Stanza: A python natural language processing toolkit for many human languages.
\newblock \emph{arXiv:2003.07082}.

\bibitem[{Ravichander et~al.(2021)Ravichander, Belinkov, and Hovy}]{ravichander_probing_2021}
Abhilasha Ravichander, Yonatan Belinkov, and Eduard Hovy. 2021.
\newblock \href {https://doi.org/10.18653/v1/2021.eacl-main.295} {Probing the probing paradigm: Does probing accuracy entail task relevance?}
\newblock \emph{ACL}, pages 3363--3377.

\bibitem[{Reimers(2019)}]{reimers2019sentence}
N~Reimers. 2019.
\newblock Sentence-bert: Sentence embeddings using siamese bert-networks.
\newblock \emph{arXiv:1908.10084}.

\bibitem[{Reimers and Gurevych(2019)}]{reimers_sentence-bert_2019}
Nils Reimers and Iryna Gurevych. 2019.
\newblock \href {http://arxiv.org/abs/1908.10084} {Sentence-{BERT}: {Sentence} {Embeddings} using {Siamese} {BERT}-{Networks}}.
\newblock ArXiv:1908.10084 [cs].

\bibitem[{Tenney et~al.(2019)Tenney, Das, and Pavlick}]{tenney_bert_2019}
Ian Tenney, Dipanjan Das, and Ellie Pavlick. 2019.
\newblock \href {https://doi.org/10.48550/arXiv.1905.05950} {{BERT} {Rediscovers} the {Classical} {NLP} {Pipeline}}.
\newblock ArXiv:1905.05950 [cs].

\bibitem[{Tesni{\`e}re(1959)}]{tesniere1959elements}
L~Tesni{\`e}re. 1959.
\newblock El{\'e}ments de syntaxe structurale.

\bibitem[{Vasileiou and Eberle(2024)}]{vasileiou-eberle-2024-explaining}
Alexandros Vasileiou and Oliver Eberle. 2024.
\newblock \href {https://doi.org/10.18653/v1/2024.naacl-long.435} {Explaining text similarity in transformer models}.
\newblock In \emph{Proceedings of the 2024 Conference of the North American Chapter of the Association for Computational Linguistics: Human Language Technologies (Volume 1: Long Papers)}, pages 7859--7873, Mexico City, Mexico. Association for Computational Linguistics.

\bibitem[{Voita and Titov(2020)}]{voita_information-theoretic_2020}
Elena Voita and Ivan Titov. 2020.
\newblock \href {https://doi.org/10.18653/v1/2020.emnlp-main.14} {Information-{Theoretic} {Probing} with {Minimum} {Description} {Length}}.
\newblock In \emph{Proceedings of the 2020 {Conference} on {Empirical} {Methods} in {Natural} {Language} {Processing} ({EMNLP})}, pages 183--196, Online. Association for Computational Linguistics.

\bibitem[{Wang et~al.(2022)Wang, Yang, Huang, Jiao, Yang, Jiang, Majumder, and Wei}]{wang2022text}
Liang Wang, Nan Yang, Xiaolong Huang, Binxing Jiao, Linjun Yang, Daxin Jiang, Rangan Majumder, and Furu Wei. 2022.
\newblock Text embeddings by weakly-supervised contrastive pre-training.
\newblock \emph{arXiv:2212.03533}.

\bibitem[{Wang et~al.(2024)Wang, Yang, Huang, Yang, Majumder, and Wei}]{wang2024multilingual}
Liang Wang, Nan Yang, Xiaolong Huang, Linjun Yang, Rangan Majumder, and Furu Wei. 2024.
\newblock Multilingual e5 text embeddings: A technical report.
\newblock \emph{arXiv:2402.05672}.

\bibitem[{Wieting et~al.(2015)Wieting, Bansal, Gimpel, and Livescu}]{wieting2015towards}
John Wieting, Mohit Bansal, Kevin Gimpel, and Karen Livescu. 2015.
\newblock Towards universal paraphrastic sentence embeddings.
\newblock \emph{arXiv:1511.08198}.

\bibitem[{Xing et~al.(2025)Xing, Luo, Xue, and Xing}]{xing2025comparativeanalysispoolingmechanisms}
Jinming Xing, Dongwen Luo, Chang Xue, and Ruilin Xing. 2025.
\newblock \href {https://arxiv.org/abs/2411.14654} {Comparative analysis of pooling mechanisms in llms: A sentiment analysis perspective}.
\newblock \emph{Preprint}, arXiv:2411.14654.

\bibitem[{Zhang et~al.(2024)Zhang, Zhang, Long, Xie, Dai, Tang, Lin, Yang, Xie, Huang, Zhang, Li, and Zhang}]{zhang-etal-2024-mgte}
Xin Zhang, Yanzhao Zhang, Dingkun Long, Wen Xie, Ziqi Dai, Jialong Tang, Huan Lin, Baosong Yang, Pengjun Xie, Fei Huang, Meishan Zhang, Wenjie Li, and Min Zhang. 2024.
\newblock \href {https://doi.org/10.18653/v1/2024.emnlp-industry.103} {{mGTE}: Generalized long-context text representation and reranking models for multilingual text retrieval}.
\newblock In \emph{Proceedings of the 2024 Conference on Empirical Methods in Natural Language Processing: Industry Track}, pages 1393--1412, Miami, Florida, US. Association for Computational Linguistics.

\bibitem[{Ács et~al.(2021)Ács, Kádár, and Kornai}]{acs_subword_2021}
Judit Ács, Ákos Kádár, and András Kornai. 2021.
\newblock \href {https://doi.org/10.48550/arXiv.2102.10864} {Subword {Pooling} {Makes} a {Difference}}.
\newblock ArXiv:2102.10864 [cs].

\end{thebibliography}

\appendix

\section{Models and Licences}

We evaluate our approach on three pre-trained sentence embedding models, all available via the HuggingFace Transformers library:

\begin{itemize}[leftmargin=*]
    \item \texttt{all-MiniLM-L6-v2} – a compact BERT-based model from the SentenceTransformers library, optimized for sentence similarity and clustering tasks. Licensed under Apache 2.0.
    
    \item \texttt{all-mpnet-base-v2} – a medium-sized MPNet model from SentenceTransformers, fine-tuned for general-purpose sentence embeddings. Licensed under Apache 2.0.
    
    \item \texttt{intfloat/multilingual-e5-large} – a multilingual encoder trained with contrastive learning across diverse languages for retrieval and semantic search tasks. Licensed under the MIT License.
\end{itemize}

All models are publicly available on HuggingFace and can be freely used for academic research.

\section{Implementation Details}

The dictionary learning code was run on Runpod on for a day and a half, with a 1 x H100 NVL with 18 vCPU 251 GB RAM. This included the full run with the models from Huggingface. 
All experiments can be reproduced by running \texttt{run\_pipeline.py} with the provided \texttt{config.yaml}. Full installation and setup instructions are provided in the README in our \href{https://github.com/matthieu-perso/mechanistic_decomposition_sentences}{Github repository}.

\section{Training and Hyperparameters}
\textbf{Probing}: We use classification (linear) and 2-layer MLPs (nonlinear) for probing token-level representations. For each classifier, we use a batch size of 128, learning rate of 1e-3, and early stopping based on validation loss. \\

\noindent \textbf{Supervised Dictionary Learning}:

We run an Optuna hyperparameter sweep for our three models \citep{optuna_2019}. We evaluate both linear and nonlinear variants of the model across embedding sizes ranging from 384 to 768, using dictionary sizes from 64 to 512 and a sparsity level of 5 non-zero coefficients. We use Adam optimizer with a dynamic learning rate. Experiments are run for 30 epochs.

\section{AI Assistants}
We used large language models (LLMs) to refine and clarify sections of the paper, including rephrasing for precision and improving stylistic consistency. All content was authored and validated by the human authors.

\section{Additional Results}
\label{sec:appendix-additional}

This appendix complements the main paper with three empirical checks:

\begin{itemize}[noitemsep, topsep=0pt, leftmargin=10pt]
  \item \textbf{Probe Results} (\autoref{sec:probe-results}): Learned atoms show near-orthogonality with a few tight clusters.
  \item \textbf{Atom Orthogonality} (\autoref{sec:atom-orthogonality}): Pairwise cosine similarities confirm separation between atoms.
  \item \textbf{POS Activation Maps} (\autoref{sec:pos-heatmaps}): Highlights model-specific encoding of syntax across the 64-atom basis.
  \item \textbf{Training Results} (\autoref{sec:training_results}): Shows dictionary convergence and training metrics.
\end{itemize}

\subsection{Probe Results}
\label{sec:probe-results}

\begin{figure}[t]
    \centering
    \includegraphics[width=\columnwidth]{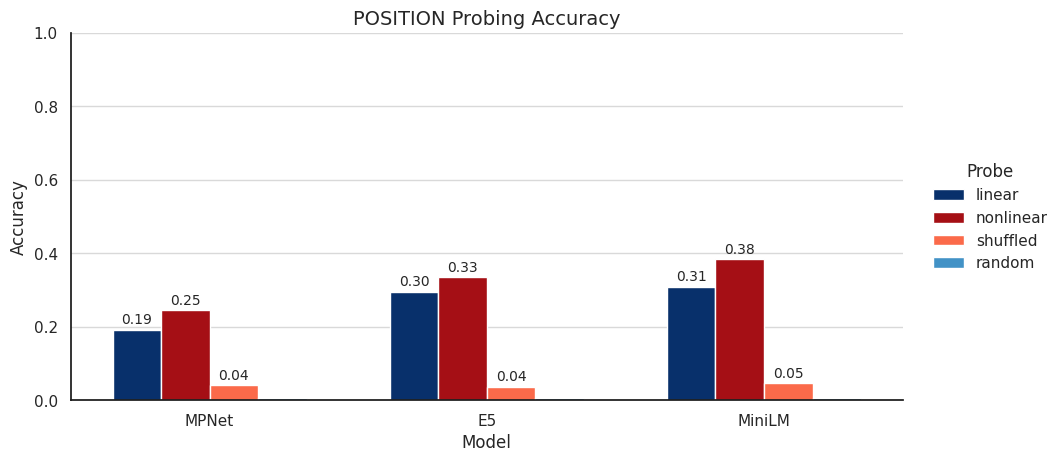}
    \caption{\small Probes for token position across models. 
} 
    \label{fig:svd_heatmap}
\end{figure}

Probes struggle to detect the position of a token in a sequence, suggesting the information is not well encoded, as shown in Figure \autoref{fig:svd_heatmap}.

\subsection{Orthogonality of Dictionary Atoms}
\label{sec:atom-orthogonality}
\begin{figure}[!htbp]
    \centering
    \includegraphics[scale=.35]{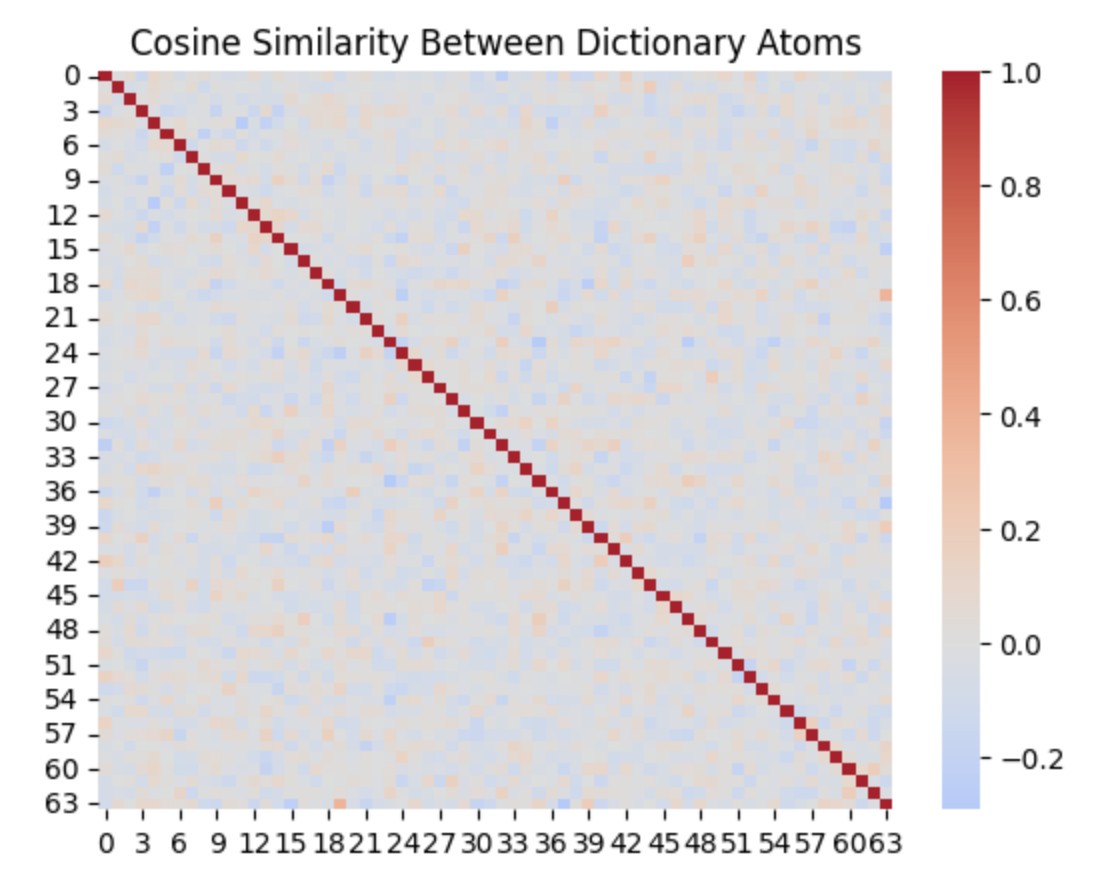}
    \caption{\small Cosine similarity of dictionary atoms, all-MiniLM
} 
    \label{fig:pca_heatmap}
\end{figure}

In \autoref{fig:pca_heatmap} we have a cosine‑similarity heatmap for the 64 MiniLM atoms. Off‑diagonal values sit near 0, showing the basis is almost orthogonal.

\begin{figure}[!htbp]
    \centering
    \includegraphics[width=\columnwidth]{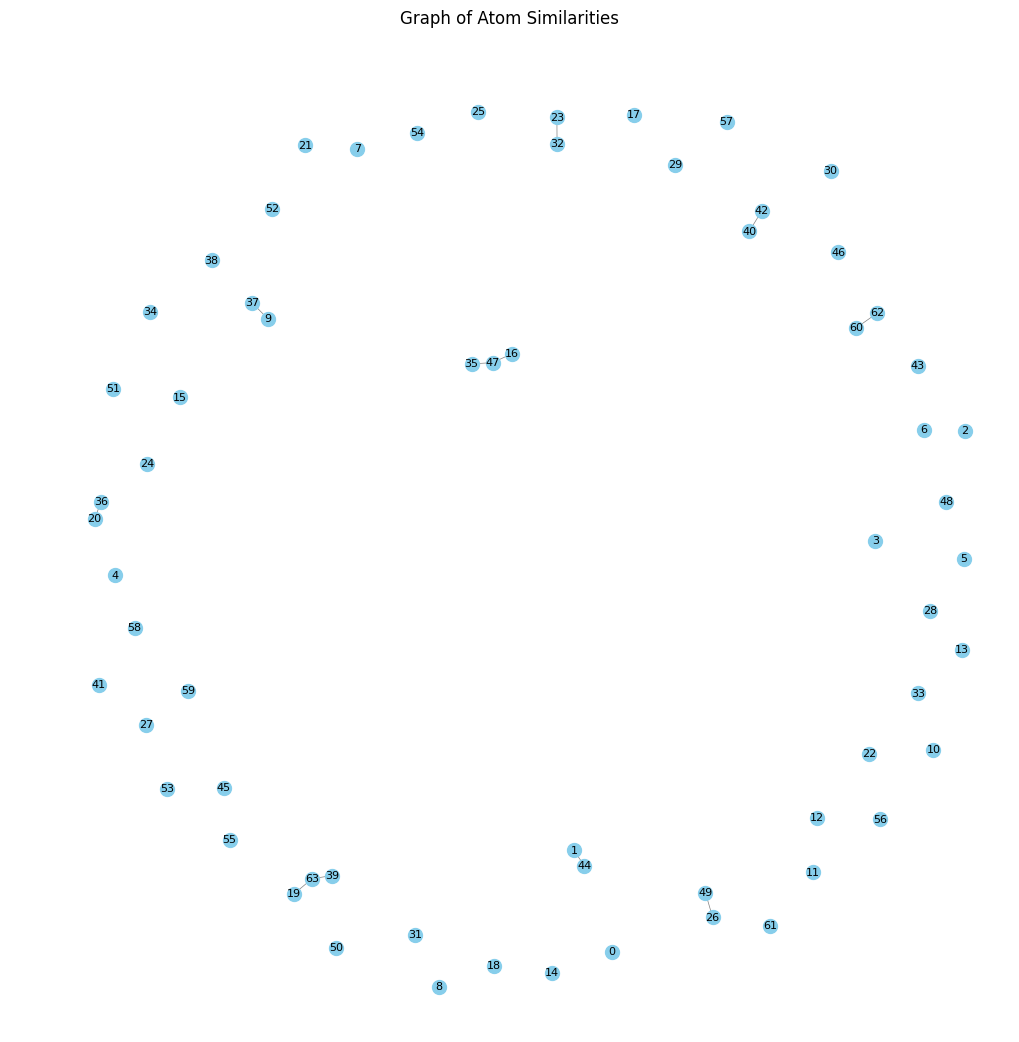}
\caption{\small Cosine similarities between dictionary atoms, for the best-performing configuration of all-MiniLM (with $k{=}64$ atoms and a linear encoder).}
    \label{fig:atom_similarities}
\end{figure}

\FloatBarrier 
\noi In \autoref{fig:atom_similarities}, the ring‑like structure and sparse edges confirm that atoms are largely independent, with only a few tight clusters (e.g.\ 35–47–16) sharing residual features.

\FloatBarrier

\subsection{Atom activiation heat map aligned with POS}\label{sec:pos-heatmaps}

\begin{figure}[!hbtp]
    \centering

    \begin{subfigure}{\columnwidth}
        \centering
        \includegraphics[width=\linewidth]{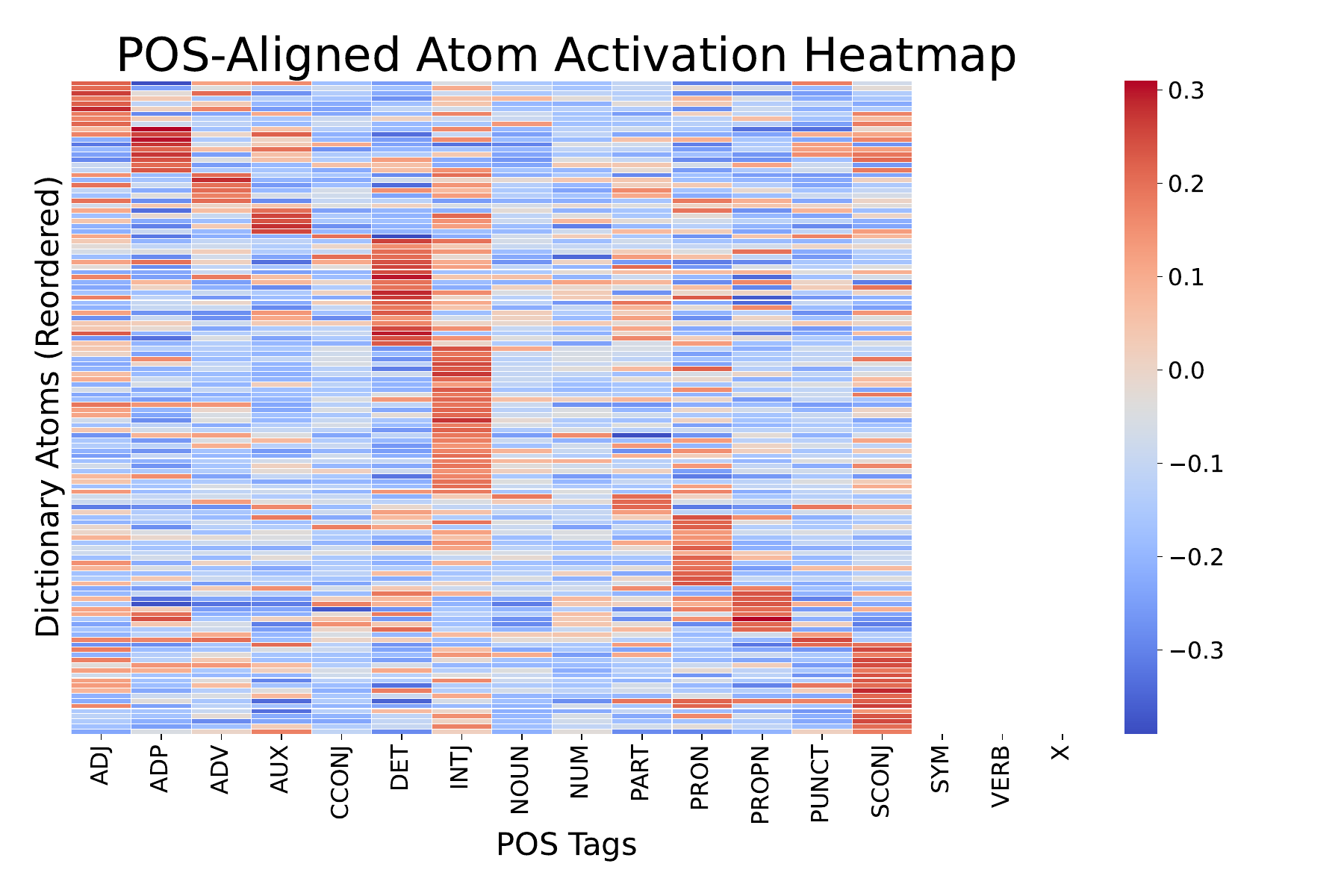}
        \caption{\small all-MiniLM-L6-v2}
    \end{subfigure}

    \begin{subfigure}{\columnwidth}
        \centering
        \includegraphics[width=\linewidth]{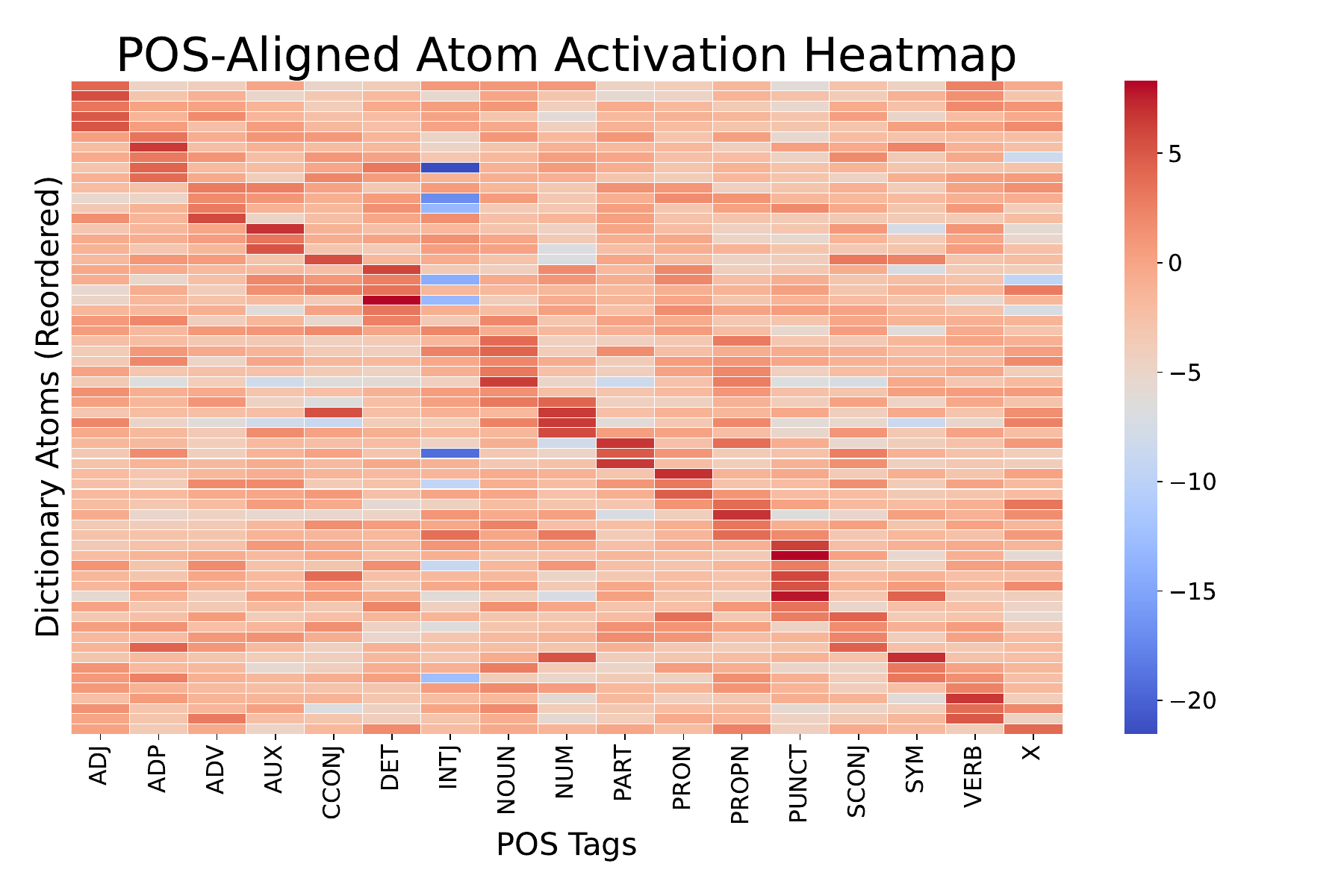}
        \caption{\small all-mpnet-base-v2 with parameters $k=64$, $lr = 0.000729$, relu, alpha-pos $= 0.827365$, aplha-dep $= 0.718469$, alpha-static $= 0.587198$, alpha-sparse $= 0.939217$, l1-s-contextual $= 0.001204$, l1-s-static $= 0.000171$}
    \end{subfigure}

    \begin{subfigure}{\columnwidth}
        \centering
        \includegraphics[width=\linewidth]{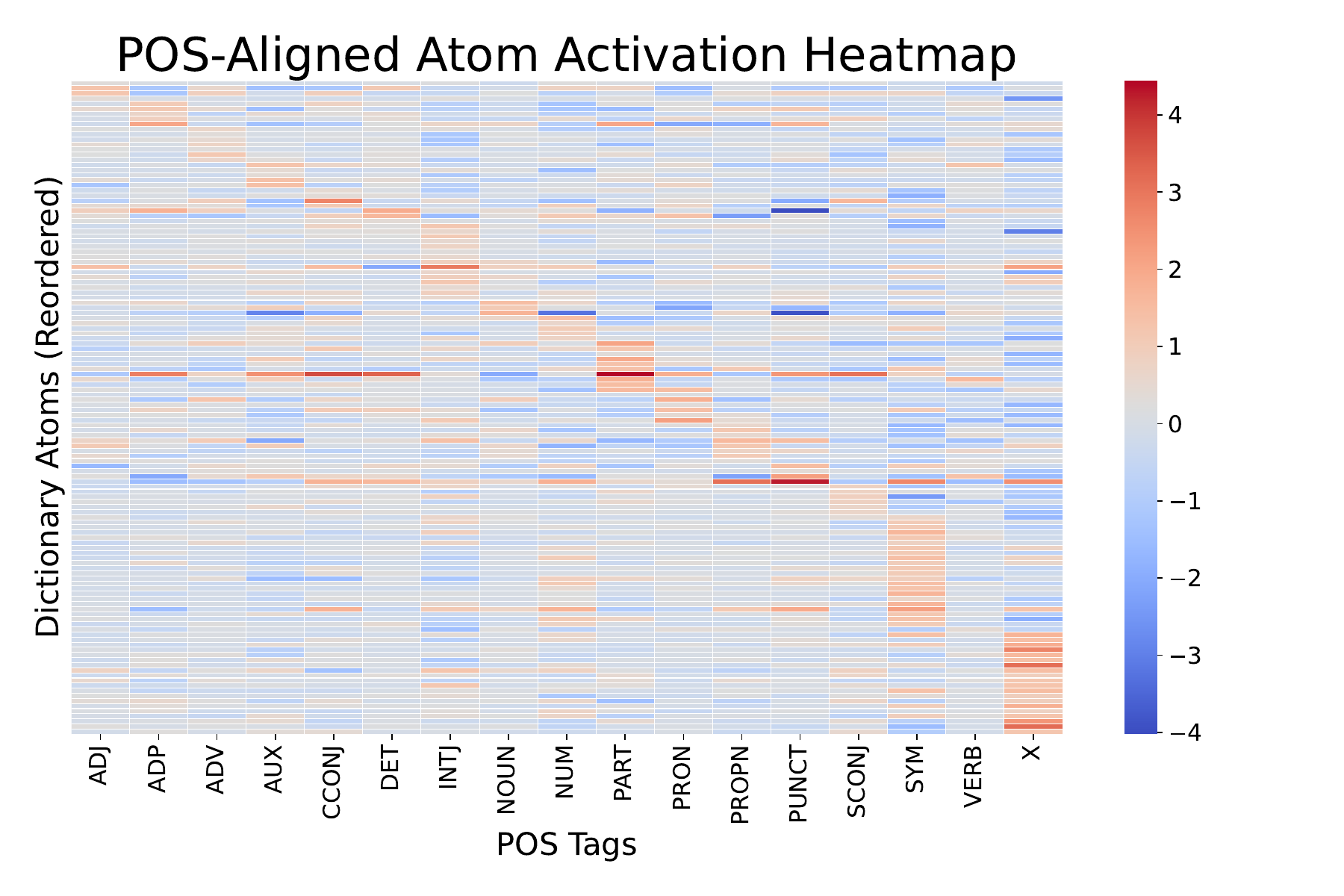}
        \caption{\small multilingual-e5-base with parameters $k=128$, $lr = 0.000443$, identity, alpha-pos $= 0.206007$, alpha-dep $=0.412214$, alpha-static $= 0.305406$, alpha-sparse $= 0.871595$, l1-s-contextual $=0.000071$, l1-s-static $= 0.0000106$}
    \end{subfigure}

    \caption{POS-aligned heatmaps for four sentence embedding models.}
    \label{fig:heatmaps-all-models}
\end{figure}

In \autoref{fig:heatmaps-all-models}, each grid shows 64 dictionary atoms (rows) versus 17 Universal POS tags (columns); 
    colours denote signed deviation from an atom’s mean activation 
    (red = stronger, blue = weaker).  
\FloatBarrier


\subsection{Training Results}
\label{sec:training_results}

\begin{table*}[t]
\centering
\resizebox{\textwidth}{!}{
\begin{tabular}{r|rrrrrrr}
\toprule
\# & \textbf{lr} & \textbf{k} & \textbf{nonlinearity} & \textbf{val\_recon} & \textbf{l1\_s\_contextual} & \textbf{f1\_pos} & \textbf{f1\_dep} \\
\midrule
1  & 0.000518 & 64  & identity & 0.120103 & 3.64e-01  & 0.021737 & 0.005369 \\
2  & 0.001739 & 64  & identity & 0.119625 & 4.34e-01  & 0.059155 & 0.035422 \\
3  & 0.000355 & 64  & identity & 0.062758 & 3.79e-04  & 0.816505 & 0.403192 \\
4  & 0.001358 & 64  & identity & 0.065688 & 1.38e-06  & 0.800894 & 0.415015 \\
5  & 0.000789 & 64  & identity & 0.066516 & 2.40e-06  & 0.805669 & 0.420882 \\
6  & 0.002610 & 64  & identity & 0.096003 & 2.96e-03  & 0.818725 & 0.276748 \\
7  & 0.000222 & 64  & identity & 0.062845 & 1.48e-05  & 0.801149 & 0.374575 \\
8  & 0.000100 & 64  & identity & 0.068060 & 4.68e-04  & 0.780040 & 0.356043 \\
9  & 0.000158 & 128 & identity & 0.060640 & 7.13e-06  & 0.740776 & 0.341501 \\
10 & 0.001912 & 128 & identity & 0.043925 & 1.74e-06  & 0.807827 & 0.438745 \\
11 & 0.005664 & 128 & identity & 0.124966 & 1.25e-02  & 0.564339 & 0.220894 \\
12 & 0.002458 & 128 & identity & 0.049120 & 3.28e-05  & 0.748126 & 0.371036 \\
13 & 0.000933 & 128 & identity & 0.033882 & 1.08e-04  & 0.784059 & 0.440293 \\
14 & 0.001061 & 128 & identity & 0.081446 & 6.30e-04  & 0.806801 & 0.393798 \\
15 & 0.006096 & 128 & relu     & 0.157965 & 1.22e-04  & 0.780371 & 0.326278 \\
16 & 0.000527 & 64  & relu     & 0.079325 & 6.47e-05  & 0.828719 & 0.434351 \\
17 & 0.000553 & 128 & relu     & 0.127369 & 1.67e-01  & 0.024133 & 0.005489 \\
18 & 0.000127 & 128 & relu     & 0.123916 & 1.71e-01  & 0.022415 & 0.005119 \\
19 & 0.004924 & 64  & relu     & 0.113595 & 3.81e-02  & 0.557816 & 0.178762 \\
20 & 0.003443 & 64  & relu     & 0.114654 & 1.63e-03  & 0.751917 & 0.357121 \\
21 & 0.001501 & 128 & relu     & 0.066949 & 5.29e-06  & 0.782448 & 0.385859 \\
22 & 0.000581 & 128 & relu     & 0.064382 & 1.08e-04  & 0.831455 & 0.427553 \\
23 & 0.000411 & 128 & relu     & 0.069176 & 2.38e-04  & 0.856974 & 0.424015 \\
24 & 0.000449 & 128 & relu     & 0.073916 & 3.80e-04  & 0.856501 & 0.407337 \\
25 & 0.000348 & 128 & relu     & 0.067931 & 4.16e-04  & 0.822681 & 0.387907 \\
26 & 0.000185 & 128 & relu     & 0.111550 & 4.47e-03  & 0.387495 & 0.151092 \\
27 & 0.000394 & 64  & relu     & 0.072853 & 7.24e-05  & 0.825091 & 0.431867 \\
28 & 0.000759 & 64  & relu     & 0.089158 & 2.40e-04  & 0.776291 & 0.434281 \\
29 & 0.000273 & 128 & relu     & 0.080254 & 1.24e-03  & 0.873602 & 0.326784 \\
30 & 0.000273 & 128 & relu     & 0.080254 & 1.24e-03  & 0.873602 & 0.326784 \\
\bottomrule
\end{tabular}
}
\caption{
Full results across hyperparameter settings for all-MiniLM, sorted by $k$ and nonlinearity. 
\texttt{val\_recon} is the mean squared error of reconstruction. 
\texttt{l1\_s\_contextual} measures sparsity of the contextual representation.
\texttt{f1\_pos} and \texttt{f1\_dep} refer to F1 scores on POS and dependency label prediction tasks, respectively.
}
\label{tab:full-results}
\end{table*}

\begin{table*}[t]
\centering
\resizebox{\textwidth}{!}{
\begin{tabular}{r|rrrrrrr}
\toprule
\# & \textbf{lr} & \textbf{k} & \textbf{nonlinearity} & \textbf{val\_recon} & \textbf{l1\_s\_contextual} & \textbf{f1\_pos} & \textbf{f1\_dep} \\
\midrule
1  & 0.005032 & 64 & identity & 0.015243 & 0.073041 & 0.410890 & 0.207144 \\
2  & 0.000854 & 64 & identity & 0.011285 & 0.004214 & 0.662754 & 0.316491 \\
3  & 0.000206 & 64 & identity & 0.012973 & 0.020591 & 0.351236 & 0.111230 \\
4  & 0.004692 & 64 & identity & 0.015362 & 0.384338 & 0.298596 & 0.107223 \\
5  & 0.005457 & 64 & relu     & 0.017434 & 0.000057 & 0.643941 & 0.281655 \\
6  & 0.000729 & 64 & relu     & 0.011658 & 0.001205 & 0.709497 & 0.344170 \\
7  & 0.001370 & 64 & relu     & 0.012290 & 0.001306 & 0.648868 & 0.362091 \\
8  & 0.000678 & 64 & relu     & 0.013526 & 0.011396 & 0.378241 & 0.157309 \\
9  & 0.000403 & 64 & relu     & 0.011158 & 0.000745 & 0.642190 & 0.352744 \\
10 & 0.003253 & 128 & identity & 0.008991 & 0.000335 & 0.690455 & 0.328152 \\
11 & 0.000556 & 128 & identity & 0.014787 & 0.209682 & 0.137597 & 0.027100 \\
12 & 0.009702 & 128 & identity & 0.015399 & 0.000442 & 0.632557 & 0.227024 \\
13 & 0.000339 & 128 & identity & 0.006979 & 0.000022 & 0.610993 & 0.292719 \\
14 & 0.003240 & 128 & identity & 0.010534 & 0.000001 & 0.582583 & 0.342678 \\
15 & 0.000520 & 128 & relu     & 0.007096 & 0.000007 & 0.617722 & 0.353173 \\
16 & 0.000775 & 128 & relu     & 0.007357 & 0.000005 & 0.690236 & 0.355175 \\
17 & 0.002351 & 128 & relu     & 0.010573 & 0.000007 & 0.638798 & 0.368472 \\
18 & 0.000115 & 128 & relu     & 0.007802 & 0.000001 & 0.593517 & 0.290650 \\
19 & 0.001707 & 128 & relu     & 0.011005 & 0.000129 & 0.669937 & 0.258801 \\
20 & 0.001843 & 128 & relu     & 0.013428 & 0.004665 & 0.587714 & 0.210425 \\
21 & 0.001060 & 128 & relu     & 0.011316 & 0.000380 & 0.573979 & 0.292974 \\
22 & 0.001112 & 128 & relu     & 0.012379 & 0.002331 & 0.706998 & 0.302391 \\
23 & 0.000297 & 128 & relu     & 0.007391 & 0.000026 & 0.702366 & 0.321286 \\
24 & 0.000235 & 128 & relu     & 0.007376 & 0.000009 & 0.598964 & 0.333313 \\
25 & 0.000104 & 128 & relu     & 0.007999 & 0.000048 & 0.552346 & 0.295003 \\
26 & 0.000524 & 128 & relu     & 0.008414 & 0.000163 & 0.669325 & 0.361527 \\
27 & 0.000492 & 128 & relu     & 0.007202 & 0.000004 & 0.624895 & 0.338907 \\
28 & 0.000288 & 128 & relu     & 0.007352 & 0.000024 & 0.660890 & 0.349720 \\
29 & 0.000728 & 128 & relu     & 0.009101 & 0.000161 & 0.691278 & 0.352695 \\
30 & 0.000690 & 128 & relu     & 0.008951 & 0.000149 & 0.693532 & 0.357377 \\
\bottomrule
\end{tabular}
}
\caption{
Full results across hyperparameter settings for MPNet-Base v2, sorted by $k$ and nonlinearity (identity first). 
\texttt{val\_recon} is the mean squared error of reconstruction. 
\texttt{l1\_s\_contextual} measures sparsity of the contextual representation.
\texttt{f1\_pos} and \texttt{f1\_dep} refer to F1 scores on POS and dependency label prediction tasks, respectively.
}
\label{tab:full-results}
\end{table*}

\begin{table*}[t]
\centering
\resizebox{\textwidth}{!}{
\begin{tabular}{r|rrrrrrr}
\toprule
\# & \textbf{lr} & \textbf{k} & \textbf{nonlinearity} & \textbf{val\_recon} & \textbf{l1\_s\_contextual} & \textbf{f1\_pos} & \textbf{f1\_dep} \\
\midrule
1  & 0.000553 & 64 & identity & 0.104789 & 0.002465 & 0.155719 & 0.060879 \\
2  & 0.001706 & 64 & identity & 0.072257 & 0.000008 & 0.782305 & 0.388732 \\
3  & 0.000267 & 64 & identity & 0.115400 & 0.082996 & 0.366908 & 0.140604 \\
4  & 0.000713 & 64 & identity & 0.063844 & 0.000118 & 0.691352 & 0.396738 \\
5  & 0.000100 & 64 & identity & 0.070072 & 0.000004 & 0.669546 & 0.297333 \\
6  & 0.000157 & 64 & identity & 0.064641 & 0.000021 & 0.756816 & 0.315536 \\
7  & 0.000657 & 64 & identity & 0.085116 & 0.003937 & 0.809701 & 0.352802 \\
8  & 0.001585 & 64 & relu     & 0.342908 & 0.011037 & 0.023774 & 0.004957 \\
9  & 0.005575 & 64 & relu     & 0.339224 & 0.002248 & 0.022222 & 0.005304 \\
10 & 0.000226 & 64 & relu     & 0.105082 & 0.001080 & 0.685338 & 0.257702 \\
11 & 0.001069 & 64 & relu     & 0.347619 & 0.121506 & 0.021822 & 0.004741 \\
12 & 0.000345 & 64 & relu     & 0.073891 & 0.000009 & 0.740304 & 0.379091 \\
13 & 0.000976 & 64 & relu     & 0.083137 & 0.000003 & 0.743548 & 0.366941 \\
14 & 0.001431 & 128 & identity & 0.046664 & 0.000019 & 0.779636 & 0.358822 \\
15 & 0.000217 & 128 & identity & 0.201939 & 0.085970 & 0.192836 & 0.047567 \\
16 & 0.005360 & 128 & identity & 0.129142 & 0.000002 & 0.712761 & 0.381553 \\
17 & 0.002346 & 128 & identity & 0.061078 & 0.000012 & 0.721825 & 0.390583 \\
18 & 0.002359 & 128 & identity & 0.064154 & 0.000039 & 0.773990 & 0.376848 \\
19 & 0.003169 & 128 & identity & 0.077541 & 0.000001 & 0.676915 & 0.414330 \\
20 & 0.009798 & 128 & identity & 0.327626 & 0.000131 & 0.683707 & 0.341344 \\
21 & 0.000444 & 128 & identity & 0.041041 & 0.000072 & 0.791771 & 0.389467 \\
22 & 0.000490 & 128 & identity & 0.081852 & 0.000241 & 0.539204 & 0.202800 \\
23 & 0.000118 & 128 & identity & 0.044689 & 0.000060 & 0.763869 & 0.376944 \\
24 & 0.000916 & 128 & identity & 0.334108 & 0.855968 & 0.023999 & 0.016383 \\
25 & 0.002350 & 128 & identity & 0.061820 & 0.000039 & 0.695598 & 0.415775 \\
26 & 0.003605 & 128 & identity & 0.121424 & 0.000425 & 0.694273 & 0.403148 \\
27 & 0.001405 & 128 & identity & 0.051058 & 0.000032 & 0.666610 & 0.383714 \\
28 & 0.001963 & 128 & identity & 0.073688 & 0.000005 & 0.707807 & 0.384857 \\
29 & 0.000478 & 128 & identity & 0.048749 & 0.000423 & 0.731864 & 0.376891 \\
30 & 0.003597 & 128 & identity & 0.097046 & 0.000077 & 0.633320 & 0.351885 \\
\bottomrule
\end{tabular}
}
\caption{
Full results across hyperparameter settings for Multilingual E5-base, sorted by $k$ and nonlinearity (identity first). 
\texttt{val\_recon} is the mean squared error of reconstruction. 
\texttt{l1\_s\_contextual} reports the L1 sparsity level of the contextual embedding.
\texttt{f1\_pos} and \texttt{f1\_dep} correspond to token-level F1 scores for part-of-speech and dependency parsing tasks, respectively.
}
\label{tab:full-results-mpnet-v2}
\end{table*}

\end{document}